\documentclass[10pt,twocolumn,letterpaper]{article}

\usepackage[pagenumbers]{cvpr} %

\usepackage{cuted}      %
\usepackage{currfile}
\usepackage{caption}
\usepackage{booktabs}
\usepackage{colortbl}   %
\definecolor{oursrow}{RGB}{232,240,251}
\usepackage{array}
\newcolumntype{L}[1]{>{\raggedright\arraybackslash}p{#1}}
\newcolumntype{N}[1]{>{\centering\arraybackslash}p{#1}}   %
\usepackage{multirow}
\usepackage{makecell}
\usepackage{enumitem}
\usepackage{xspace}
\usepackage{siunitx}
\newcommand{\TS}{TotalSegmentator\xspace}
\newcommand{\MOOSE}{MOOSE\xspace}
\newcommand{\nnunet}{nnU-Net\xspace}
\newcommand{\pgps}{PGPS\xspace}
\newcommand{\enh}{ENHANCE.PET\xspace}
\newcommand{\Dsc}{DSC\xspace}

\definecolor{cvprblue}{rgb}{0.21,0.49,0.74}
\usepackage[pagebackref,breaklinks,colorlinks,allcolors=cvprblue]{hyperref}
\usepackage{xurl}

\newcommand{\codeavailabilityabstract}{We provide the method as a drop-in for nnU-Net, TotalSegmentator and MOOSE. The
code is part of nnU-Net (\url{https://github.com/MIC-DKFZ/nnUNet}), and unified models for both toolkits will be released.}
\newcommand{\codeavailabilityintro}{Our predictor and trainers are part of
\nnunet.\footnote{\url{https://github.com/MIC-DKFZ/nnUNet}. On the master branch, and part of \nnunet from version
2.9.0.} The integrations into TotalSegmentator and MOOSE
(Sec.~\ref{sec:release}) are available as pull requests to both toolkits.\footnote{\url{https://github.com/wasserth/TotalSegmentator/pull/648}
and \url{https://github.com/ENHANCE-PET/MOOSE/pull/250}} The two unified models of Sec.~\ref{sec:onemodel} are available
at temporary links.\footnote{TotalSegmentator classes: \url{https://hub.dkfz.de/s/AYdkSQ75Yysqsid}; MOOSE classes:
\url{https://hub.dkfz.de/s/7Af5nRjiLTw7SHb}.} They will be replaced by different models, which will also be the official
releases of TotalSegmentator and MOOSE.}

\title{Segment Everything, Materialize Nothing:\\Exact Streaming Inference for Many-Class 3D Segmentation}

\author{\parbox{\textwidth}{\centering
  \mbox{Fabian Isensee$^{1,2}$}\quad
  \mbox{Jakob Wasserthal$^{3}$}\quad
  \mbox{Lalith Kumar Shiyam Sundar$^{4}$}\quad
  \mbox{Maximilian Rokuss$^{1,5}$}\quad
  \mbox{Moritz Langenberg$^{1,5}$}\quad
  \mbox{Constantin Ulrich Harsy$^{1,6}$}\quad
  \mbox{Stefan Michael Happold$^{7}$}\quad
  \mbox{Klaus H. Maier-Hein$^{1,2,5,6,8,9}$} \\[6pt]
  {\small
  $^1$Division of Medical Image Computing (MIC), German Cancer Research Center (DKFZ), Heidelberg, Germany \\
  $^2$National Center for Tumor Diseases (NCT) Heidelberg, a partnership between the German Cancer Research Center (DKFZ) and
  Heidelberg University Hospital (UKHD), Heidelberg, Germany \\
  $^3$Clinic of Radiology and Nuclear Medicine, University Hospital Basel, Basel, Switzerland \\
  $^4$DIGITX Lab, Department of Radiology, LMU University Hospital, Munich, Germany \\
  $^5$Faculty of Mathematics and Computer Science, Heidelberg University, Germany \\
  $^6$Helmholtz Imaging, German Cancer Research Center (DKFZ), Heidelberg, Germany \\
  $^7$School of Computation, Information and Technology, Technical University Munich, Munich, Germany \\
  $^8$Pattern Analysis and Learning Group, Department of Radiation Oncology, Heidelberg University Hospital, Heidelberg, Germany \\
  $^9$Medical Faculty, Heidelberg University, Germany \\[2pt]
  \texttt{f.isensee@dkfz-heidelberg.de}}}}

\begin{document}
\maketitle
\begin{abstract}
Whole-body segmentation tools such as TotalSegmentator and MOOSE split more than 100 classes across five or six networks. The
reason is memory. Standard 3D inference, as implemented in nnU-Net and MONAI, holds a logit for every class and voxel, on the
working grid and again on the scanner's grid. For a median whole-body CT at 125 classes, that is 76\,GB (Fig.~1). We show that
this memory wall lies in what inference materializes, not in the task, and remove it without approximation. Every step after the
network is local along one axis, so its outputs can be streamed through a ring one patch thick. The classification head commutes
with blending and resampling, so the ring carries the network's features instead of the class logits. Memory thus no longer grows
with classes times volume, and the label map on the scanner's grid stays exact. This holds for any model with a pointwise affine
head, without retraining. Many-class models, previously impractical, become deployable. One model trained on all classes matches or
slightly exceeds the toolkits' five or six specialists in segmentation quality and runs 5.8--15$\times$ faster end to end, with
3.2--7.6$\times$ less RAM and under 8\,GB of VRAM. \codeavailabilityabstract
\end{abstract}

\begin{figure}[t]
\centering
\includegraphics[width=\linewidth]{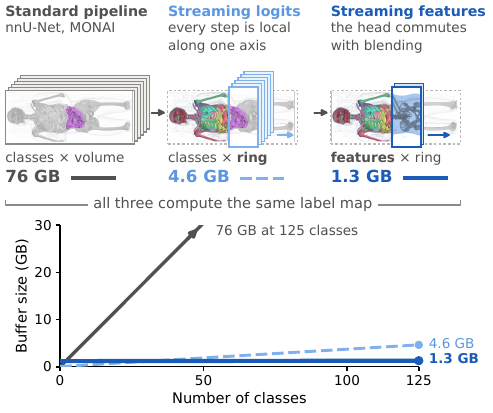}
\caption{\textbf{Many-class 3D segmentation does not need image-sized logits.} Top: the standard (canonical) pipeline holds logits for every class over the whole volume (left). Every step after the network is local along one axis, so the logits can be streamed through a ring of slices (middle). The classification head commutes with blending, so the ring can hold features instead of logits (right). All three compute the same label map in exact arithmetic. Bottom: buffer size vs.\ the number of classes; only the feature ring stays flat. Sizes for a median \enh CT, at 125 classes.}
\label{fig:teaser}
\end{figure}

\section{Introduction}
\label{sec:intro}
Label spaces in 3D medical image segmentation keep growing as datasets are merged and refined.
TotalSegmentator~\cite{wasserthal2023totalsegmentator,wasserthal2026tsv3data} delineates 117 structures in CT, MOOSE~\cite{shiyamsundar2022moose} 125 on
the \enh data~\cite{ferrara2026enhance}, and whole-body label sets with more than 140 classes exist~\cite{jaus2023atlas,xu2025cads}.
Yet both tools split their label space across five or six separately trained networks. The reason is the inference pipeline that
the field relies on, implemented in widely used frameworks such as \nnunet~\cite{isensee2021nnu} and MONAI~\cite{cardoso2022monai}.
It first resamples each scan to a common working spacing, so that object sizes are consistent across scans. It predicts logits
(per-class outputs before softmax) on overlapping windows and blends them with Gaussian weights. It then maps the image-sized logits
back to the scanner's grid and takes the argmax there, which places boundaries as exactly as possible.

At many classes, this pipeline is limited by a single tensor, the logits per class and voxel at image size. For a median
whole-body CT of \enh (512$\times$512$\times$381 voxels) at 125 classes, these logits occupy 46.5\,GB on the scanner's grid and
another 29.7\,GB at the working resolution. A 2D image of the same in-plane size would need 0.12\,GB. It is the third dimension
that turns the number of classes into a memory problem. Canonical inference at these class counts is therefore out of reach for
end users such as clinicians, who have consumer GPUs or none at all (Fig.~\ref{fig:teaser}).

TotalSegmentator and MOOSE (from here on, the toolkits) work around this tensor instead of removing it. Each of their five or six
networks covers a subset of the classes, which keeps its logits small enough to handle, but every scan now passes through all of
them, each with its own preprocessing and export. Even so, one network's logits are too large to resample. The toolkits therefore
take the argmax at the working resolution and upsample the label map with nearest neighbor. This puts every boundary on the working
grid, which produces staircase contours (Fig.~\ref{fig:blocks}) and lowers the Dice similarity coefficient (\Dsc) against
references drawn on the scanner's grid.

We show that this memory wall lies in what the pipeline materializes, a logit tensor whose size is the number of classes times the
volume, and not in the task. Two observations remove one factor each. First, every step after the network is local along one axis.
The network predicts one patch at a time, a blended slice depends only on the windows that cover it, the head and the argmax act per
voxel, and linear resampling needs two neighboring slices. A ring buffer one patch thick therefore suffices, and finished slices are
resampled and reduced to labels on the scanner's grid. On the median CT at 125 classes, this shrinks the logits from 76 to 4.6\,GB
(Fig.~\ref{fig:teaser}), but the ring still holds every class. Second, a segmentation network computes its $C$ logits at every voxel
from $F<C$ decoder features through a pointwise affine head ($F=32$ in \nnunet's U-Nets, $C=117$--125 here). Blending, test-time
mirroring and linear resampling are weighted averages, so the head commutes with them, and the ring can hold $F$ features instead
of $C$ logits (1.3\,GB). Only the argmax, the one nonlinear step, comes last. Together, the two observations compute the label map
of the canonical pipeline up to floating-point rounding, and the class count never multiplies the volume.

The second observation matters already at today's class counts. At 125 classes it shrinks the ring from 4.6 to 1.3\,GB, which
decides whether inference, network included, fits on an 8\,GB GPU (Sec.~\ref{sec:combined}). A ring of logits would also keep
growing with every class.

Exact streaming inference removes a constraint that the field has worked around for years, with two consequences. First, every \nnunet model, existing
or new, gets exact many-class inference as a drop-in predictor, because streaming changes only how the output is computed. Second,
unified many-class models, previously practical only at reduced resolution and with nearest-neighbor upsampling, become deployable.
In TotalSegmentator and MOOSE, one model per label space replaces the five or six networks with matching or slightly better
segmentation quality. End to end, one 117-class model segments a whole-body CT on the scanner's grid in 5.8--11\,s with under
7\,GB of RAM and 8\,GB of VRAM, 12--15$\times$ faster than the released TotalSegmentator, and one 125-class model is
5.8--7.6$\times$ faster than the released MOOSE (Sec.~\ref{sec:release}).

\noindent\textbf{Contributions.}
\begin{itemize}[leftmargin=*,itemsep=1pt,topsep=2pt]
  \item \textbf{Where the memory goes.} An analysis showing that the memory wall of many-class 3D inference lies in what the
  canonical pipeline materializes, not in the task. The pipeline holds $C$ logits at image size on two grids,
  $O(C(N+M))$ (Sec.~\ref{sec:pipeline}).
  \item \textbf{Exact streaming inference.} A method built on two observations. First, every step after the network is local along
  one axis, so the logits can be streamed instead of being held for the whole volume (Sec.~\ref{sec:streaming}). Second, the affine
  head commutes with blending, mirroring and resampling (Proposition~1), so the stream can carry $F$ features instead of $C$ logits
  (Sec.~\ref{sec:reorder}). The label map stays exact for any model with a pointwise affine head, and the class count never
  multiplies the volume.
  \item \textbf{Deployable many-class segmentation.} Every existing \nnunet model gets exact many-class inference without
  retraining. One model replaces TotalSegmentator's five and MOOSE's six specialist models with matching or slightly better segmentation quality, at a
  fraction of their inference time and memory (Secs.~\ref{sec:onemodel},~\ref{sec:release}).
\end{itemize}
\codeavailabilityintro

\section{Related work}
\label{sec:related}

\paragraph{Whole-body segmentation toolkits.}
Whole-body segmentation toolkits cover large label spaces by splitting them across several networks to manage memory.
TotalSegmentator~\cite{wasserthal2023totalsegmentator,wasserthal2026tsv3data} segments 117 structures in CT with five networks at 1.5\,mm and offers a
single all-class model only at 3\,mm and 6\,mm. In its release, TotalSegmentator-MRI~\cite{dantonoli2025totalsegmentatormri}
segments 50 structures in MRI with two networks. MOOSE~\cite{shiyamsundar2022moose} segments 125 anatomical structures of the \enh PET/CT
dataset~\cite{ferrara2026enhance} with six networks at 1.5\,mm. CADS~\cite{xu2025cads} segments 167 structures in CT with nine networks at 1.5\,mm. All four build on \nnunet~\cite{isensee2021nnu} and limit memory in two more ways. They hold the logits only at the working
resolution and take the argmax there, and they bring the label map to the scanner's grid by nearest-neighbor upsampling, an
approximation of the exact export. We focus on
TotalSegmentator and MOOSE, the most widely used of these toolkits, keep their label spaces and training data, and remove the
reason for the split.

\paragraph{Memory in sliding-window inference.}
Common sliding-window implementations, including those of \nnunet, MONAI and TorchIO~\cite{perezgarcia2021torchio}, accumulate
the logits of all windows in one image-sized buffer, so the class dimension is held at image size. In 2D, mask-classification
models~\cite{cheng2021maskformer,cheng2022mask2former} predict a fixed set of masks and classify each of them, so their mask
output does not grow with the number of classes. They are a
different model family, whereas our reordering applies unchanged to the per-voxel classifiers of 3D medical segmentation.

\paragraph{Low-rank heads.}
Closest to our work, Jaubert et al.~\cite{jaubert2026breaking} replace the head of an \nnunet model with a trained low-rank head,
which maps to a few basis channels and projects them to classes. This allows chunking the training loss over classes. At
inference, they blend the basis maps instead of the logits, and their pipeline ends at the working resolution. Our approach
differs in three respects. First, it streams, so no volume-sized intermediate is held, neither logits nor features. Second, it
keeps the network's own head, so existing models need no retraining. Blending its $F$ input features is exact for any affine
head, because blending is a weighted average (Proposition~1). Third, it exports exactly on
the scanner's grid, where the larger class-sized tensor lives (46.5 vs.\ 29.7\,GB).

\paragraph{Universal and promptable models.}
Universal models condition one network on the requested classes, through learned class embeddings in VISTA3D~\cite{he2025vista3d} (127 structures, built on MONAI~\cite{cardoso2022monai}) or text prompts in SAT~\cite{zhao2023sat} and VoxTell~\cite{rokuss2026voxtell}. At inference, however, they still produce one logit map per requested
class and therefore hold the class-sized tensor of Sec.~\ref{sec:pipeline}. VISTA3D blends these maps at image size and upsamples
its label map with nearest neighbor, and VoxTell holds one map per prompt on the scanner's grid. Sec.~\ref{sec:toolkits}
measures what this costs at the class counts of TotalSegmentator and MOOSE. %

\paragraph{Memory-efficient exact computation.}
Avoiding the materialization of large intermediate tensors is an established principle in deep learning.
FlashAttention~\cite{dao2022flashattention} reorders and tiles attention so that the attention matrix is never materialized, and cut
cross-entropy~\cite{wijmans2025cce} computes the loss over large vocabularies without materializing the logits. We apply the same
principle to the inference of dense prediction, with an exact result and no logits at image size.

\begin{figure*}[t]
\centering
\includegraphics[height=1.75in]{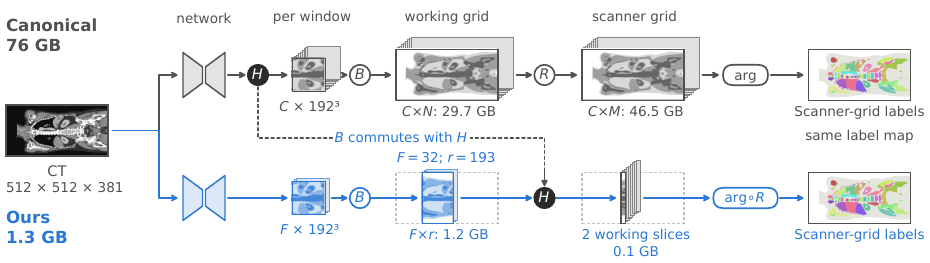}
\caption{\textbf{Our pipeline computes the same label map as the canonical one, with 1.3 instead of 76\,GB of buffers.} Example: the median \enh CT of Fig.~\ref{fig:teaser}, with $C=125$ and $F=32$. Top row: canonical inference applies the affine head $H$ per window; blending $B$ (Gaussian-weighted averaging) and linear resampling $R$ hold $C$ logits per voxel on the working ($N$) and scanner ($M$) grids. Bottom row: ours blends $F$ features in a ring (here $r=193$), applies $H$ per completed slice, and uses two slices of logits for $R$, then argmax, fused. Dashed arrow: $B$ commutes with $H$ (Proposition~1). Stack depth denotes channels. Equivalence holds in exact arithmetic (Sec.~\ref{sec:combined}).}
\label{fig:streaming}
\end{figure*}

\section{Method}
\label{sec:method}

\subsection{The canonical pipeline and its memory cost}
\label{sec:pipeline}
A segmentation network maps an image window to decoder features $f(x)\in\mathbb{R}^{F}$ at every voxel $x$ of the working grid.
A pointwise affine head $H$ then computes the logits $z(x) = Wf(x)+b$, with $W\in\mathbb{R}^{C\times F}$. In sliding-window
inference, the logits of overlapping windows $w$ are blended with non-negative weights $g_w$ (Gaussian, zero outside the window).
This blending $B$ computes $Z(x) = \sum_w g_w(x)\,z_w(x) / \sum_w g_w(x)$, and test-time mirroring averages the logits of flipped
windows in the same way. The blended logits are then resampled ($R$) to the voxels $y$ of the scanner's grid,
$\hat Z(y) = \sum_x \rho_{yx} Z(x)$, and the argmax gives the label $\hat\ell(y) = \arg\max_c \hat Z_c(y)$. In operator form (Fig.~\ref{fig:streaming}, top row),
\begin{equation}
  \hat\ell = \operatorname{arg}\circ R\circ B\circ H\,(f).
  \label{eq:canonical}
\end{equation}
Inside the network, the number of classes only enters the head ($CF$ weights), so the cost of $C$ lies in the tensors around it.
$Z$ and $\hat Z$ have $C$ channels on the $N$ voxels of the working grid and the $M$ voxels of the scanner's grid, $O(C(N+M))$ in
total. For the median whole-body CT of Sec.~\ref{sec:intro} at 125 classes ($N=317{\times}317{\times}635$ voxels at 1.5\,mm,
$M=512{\times}512{\times}381$), these take 29.7\,GB and 46.5\,GB. To avoid the larger tensor $\hat Z$, the toolkits
take the argmax on the working grid and upsample the label map with nearest neighbor,
$\hat\ell_\text{NN} = R_\text{NN}\circ\operatorname{arg}\circ B\circ H\,(f)$. $Z$ is still held, and every boundary lies on the
working grid.

\subsection{Spatial streaming}
\label{sec:streaming}
Our method keeps the network and its head unchanged and reorganizes only the steps after them.
Every step after the network is local along one axis. The network predicts one patch at a time, and a blended slice depends only
on the windows that cover it. The head and the argmax act per voxel, and linear resampling needs two neighboring slices.
The blended volume therefore never has to exist as a whole, and we stream along one axis of the working grid instead
(Fig.~\ref{fig:streaming}). Streaming processes the windows slab by slab along this axis and blends their predictions into a ring buffer. Once all windows
of a slab are predicted, the part of the slab that no future window touches is complete and is exported, and the slab advances
by the sliding-window step. Resampling also needs the last completed slice, so the ring holds $r=193$
slices for a 192$^3$ patch. With the head applied per window, as in the canonical order, the ring holds $C$ channels. With $S$ voxels per
slice, the buffers shrink from $O(C(N+M))$ to $O(CrS)$, 4.6 instead of 76\,GB for the median whole-body CT at
125 classes. They still grow with the class count, which the next step removes.

\subsection{Algebraic reordering}
\label{sec:reorder}
\noindent\textbf{Proposition 1.} \emph{Let $A$ be a linear operator that acts on every channel alike, $(Au)(y)=\sum_x a_{yx}u(x)$,
with $\sum_x a_{yx}=1$ for every $y$. Then $A$ commutes with the head: $A(Wf+b) = W(Af)+b$.}

\noindent The proof is linearity: $\sum_x a_{yx}(Wf(x)+b) = W\sum_x a_{yx}f(x) + b\sum_x a_{yx} = W(Af)(y) + b$, since the
weights sum to one. Blending, mirroring and linear interpolation are such operators, and so is any composition of them. B-spline
interpolation qualifies as well, as its weights also sum to one. The identity is elementary. Still, common sliding-window
implementations apply the head per window and blend the logits (Sec.~\ref{sec:related}).
By Proposition~1,
\begin{equation}
\begin{aligned}
  \hat\ell \;&=\; \operatorname{arg}\circ R\circ B\circ H\,(f)\\
  &=\; \operatorname{arg}\circ R\circ H\circ B\,(f) \;=\; \operatorname{arg}\circ H\circ R\circ B\,(f):
\end{aligned}
  \label{eq:reorder}
\end{equation}
the head may act at any point before the argmax (dashed arrow in Fig.~\ref{fig:streaming}). A pipeline can therefore blend and resample $F$ features instead of $C$ logits
and obtain the same label map in exact arithmetic. Put differently, every logit vector lies in the affine subspace $\{Wv+b: v\in\mathbb{R}^F\}$ of
$\mathbb{R}^C$, of dimension at most $F$. Only the argmax cannot move, because it is not affine and does not commute
with $R$. This is why the toolkits' $\hat\ell_\text{NN}$ is an approximation (Sec.~\ref{sec:nn}).

\paragraph{Scope.} Proposition~1 only requires a fixed pointwise affine head. Convolutional and transformer U-Nets end in such a
head, a $1{\times}1{\times}1$ convolution, and so does VISTA3D's automatic mode. The blending must act on the logits, as it does in
\nnunet and MONAI, and not on per-window probabilities. Per-voxel thresholds of sigmoid outputs are covered in the same way as the
argmax. Ensembles of $K$ models are covered as well. Their ring holds the members' $KF$ features, or their averaged $C$ logits where these are fewer (Suppl.~S10).
Decoders that take the class as an input, such as VoxTell's, have no such head. Their output can still be streamed, but with
$C$ channels in the ring (Sec.~\ref{sec:streaming}).

\subsection{Exact streaming inference}
\label{sec:combined}
\begin{table*}[t]
\caption{\textbf{Exact streaming inference is faster and needs far less RAM than all other pipelines.} With the toolkits' five or six
networks it is 1.8--2.3$\times$ faster. One network for all classes, run with our inference, is 6.3--13$\times$ faster than the releases. \nnunet 2.8.1 and MONAI run
out of RAM. Numbers show the prediction step only, from the preprocessed image to the label map on the scanner's grid (for end to
end, see Tab.~\ref{tab:release}). Four \enh test scans, one per site, same settings for every pipeline (Sec.~\ref{sec:experiments}); ranges
over the scans, per scan in Suppl.~Tab.~S1. Plain U-Net: the specialists' network. ResEnc U-Net: our model. Slower than ours: time
relative to ours with the plain or the ResEnc U-Net, per scan. Toolkit pipeline: the toolkit's own inference on one network
for all classes. $^\ddagger$Out of RAM on one scan; ranges over the other three.}
\label{tab:toolkits}
\centering
\footnotesize
\setlength{\tabcolsep}{3.4pt}
\begin{tabular}{@{}lccccc@{\hspace{1.4em}}ccccc@{}}
\toprule
 & \multicolumn{5}{c}{\textbf{\TS labels (117 classes)}} & \multicolumn{5}{c}{\textbf{\enh labels (125 classes, MOOSE)}} \\
\cmidrule(r{1.4em}){2-6}\cmidrule{7-11}
 & time & RAM & VRAM & \multicolumn{2}{c}{slower than ours} & time & RAM & VRAM & \multicolumn{2}{c}{slower than ours} \\
\cmidrule(lr{1.4em}){5-6}\cmidrule(l){10-11}
pipeline & (s) & (GB) & (GB) & plain & ResEnc & (s) & (GB) & (GB) & plain & ResEnc \\
\midrule
\multicolumn{11}{@{}l}{\textit{5 / 6 plain U-Nets, the toolkit's split}} \\
\quad release & 33--95 & 8.2--21 & 9.6--21 & 8.7--9.7$\times$ & 6.3--8.8$\times$
  & 46--106 & 5.6--14 & 4.6--8.5 & 10--13$\times$ & 8.6--9.6$\times$ \\
\rowcolor{oursrow} \quad Ours & 17--52 & 2.0--5.1 & 2.5--3.3 & 4.7--4.9$\times$ & 3.1--4.8$\times$
  & 22--49 & 2.1--5.7 & 3.0--4.1 & 5.5--5.8$\times$ & 4.1--5.1$\times$ \\
\multicolumn{11}{@{}l}{\textit{1 plain U-Net, all classes}} \\
\quad toolkit pipeline & 50--104$^\ddagger$ & 33--53$^\ddagger$ & 15--23$^\ddagger$ & 14--17$\times^\ddagger$ & 9.5--14$\times^\ddagger$
  & 25--121 & 10--42 & 12--22 & 6.3--15$\times$ & 4.7--11$\times$ \\
\quad \nnunet 2.8.1, MONAI & \multicolumn{5}{c}{\textit{out of RAM on every scan}} & \multicolumn{5}{c}{\textit{out of RAM on every scan}} \\
\rowcolor{oursrow} \quad Ours & 3.4--11 & 1.7--3.2 & 2.7--3.8 & 1 & 
  & 4.0--8.9 & 1.7--2.0 & 3.3--4.4 & 1 & \\
\multicolumn{11}{@{}l}{\textit{1 ResEnc U-Net, all classes}} \\
\rowcolor{oursrow} \quad Ours & 5.3--12 & 2.1--3.4 & 6.1--7.3 & 1.0--1.5$\times$ & 1 & 5.3--12 & 2.1--2.5 & 6.1--7.3 & 1.1--1.3$\times$ & 1 \\
\bottomrule
\end{tabular}
\end{table*}
Combined, the ring of Sec.~\ref{sec:streaming} holds the blended $F$ features instead of $C$ logits. We compute the middle form of
Eq.~\eqref{eq:reorder} (Fig.~\ref{fig:streaming}, bottom row): the head is applied once to every finished working slice, and its $C$ logits are kept for the next output
plane. A fused GPU kernel resamples the two working slices around each output plane, first in-plane and then along the axis, and keeps a running maximum over the classes, writing the label directly. Applying the head before the resampling is the cheaper order, because a working slice
has fewer voxels than the output planes it serves; Eq.~\eqref{eq:reorder} makes either order exact. Our predictor streams the
canonical order only when $C\le F$, where its ring is the smaller one.

\paragraph{Memory.} The logits take $O(C(N+M))$ in the canonical pipeline, $O(CrS)$ when streamed (Sec.~\ref{sec:streaming})
and $O((Fr+2C)S)$ in ours, plus the label map at one or two bytes per voxel. In our pipeline, the number of classes therefore
multiplies only two slices of logits and never the volume. For the median whole-body CT at 125 classes, the three pipelines need
76, 4.6 and 1.3\,GB (Fig.~\ref{fig:teaser}), of which our feature ring takes 1.2\,GB and the logits of two slices 0.1\,GB. The gap
grows with every class. On this scan, each additional class adds 0.6\,GB to the canonical tensors, 38\,MB to the streamed logits
and 0.8\,MB to ours. At a hypothetical 1000 classes, this amounts to 610, 37 and 1.9\,GB, and the streamed logits alone would
exceed what consumer GPUs hold. The remaining class-dependent term, the logits of two slices, could be removed by applying the head
to a few classes at a time and keeping a running maximum over them. This would save only 0.1\,GB at 125 classes (0.75\,GB at
1000) at the cost of more passes over the two slices, so we do not do it.

\paragraph{Measured VRAM.} We measured the full VRAM requirement of our predictor, including the network's forward pass, on the
four scans of Tab.~\ref{tab:toolkits}, with the network compiled (Tab.~\ref{tab:toolkits} runs it eager; Suppl.~S10). It is
5.4--6.5\,GB with the feature ring and 8.8--12.8\,GB with a ring of logits, and the feature ring needs 11--14\,\% less time. This
decides whether an 8\,GB GPU suffices.

\paragraph{Floating point.} In exact arithmetic, streaming and the canonical pipeline produce the same label map. With the network
in fp32, they agree up to 0--3 voxels per scan, all at near-ties of the two highest logits. By default, \nnunet runs networks in mixed precision. Our label maps are then closer to a full-fp32 reference than even those of the
canonical pipeline, because we apply the head in fp32 after blending. They differ from it in 0.12 and 0.050\,\% of the foreground
on \TS and \enh, against 0.20 and 0.096\,\% for the canonical pipeline (Suppl.~S10).

\section{Training unified many-class models}
\label{sec:training}

With streaming inference, one model can cover a whole label space. \nnunet 2.8.1 cannot train a model with 117 classes and a 192$^3$ patch size on a 48\,GB GPU. Its per-epoch validation runs out of memory from 96 classes onwards. Even if it could, training
would be held back in two more places. Spatial augmentation interpolates the label map linearly as $C$ one-hot channels, which
takes 4.6\,s of CPU time per sample. The deep-supervised loss materializes softmax probabilities, one-hot targets
and Dice terms at every output scale, which costs 6.3\,GB of VRAM. We therefore replace the one-hot predictions of the per-epoch validation with a $C{\times}C$ confusion matrix, resample the
label map with nearest neighbor (0.4--0.6\,s per sample), and compute the loss in fused kernels that do not store these tensors. With these changes, a 117-class
model trains in 30\,GB of VRAM, and its training step is 12\,\% faster than that of \nnunet 2.8.1 (Suppl.~S7).

\paragraph{Schedule.}
Unified models benefit from longer schedules than \nnunet's default of 1000 epochs (Suppl.~S9), which makes training expensive.
We therefore use progressive growing of patch size (\pgps)~\cite{fischer2024pgps,fischer2026pgps}. It starts training on small
patches and grows them to the full patch size as training progresses. All our models are trained with \pgps over 5400 epochs,
chosen by cross-validation, at 1.6$\times$ the cost of 1000 dense epochs. One ResEnc U-Net at 192$^3$ takes 79 and 80 L40S
GPU-hours on \TS and \enh. The toolkits' five and six specialists, plain U-Nets with smaller patch sizes trained for 1000 or 2000 epochs, take a total of 42 and 102 GPU-hours.

\section{Experiments}
\label{sec:experiments}

\begin{table}[t]
\caption{\textbf{One ResEnc U-Net per label space matches or slightly exceeds five or six specialists.} Mean \Dsc over all classes on the test
splits (n=109 / 337), all models exported exactly. Specialists: the toolkits' five or six networks, retrained with their recipes on the same
training splits. Plain U-Net: the specialists' network used in our training recipe
(Sec.~\ref{sec:training}). ResEnc U-Net: our model. FP-penalized: a class predicted in a scan that does not contain it scores 0; a class absent from both reference and
prediction is not scored. Present structures: only classes present in the reference are scored. $\Delta$: difference to the specialists; $^*$: its 95\,\% paired
case-bootstrap interval excludes 0.}
\label{tab:onemodel}
\centering
\footnotesize
\setlength{\tabcolsep}{5pt}
\begin{tabular}{@{}lcccc@{}}
\toprule
 & \multicolumn{2}{c}{FP-penalized} & \multicolumn{2}{c}{present structures} \\
\cmidrule(lr){2-3}\cmidrule(l){4-5}
 & \Dsc & $\Delta$ & \Dsc & $\Delta$ \\
\midrule
\multicolumn{5}{@{}l}{\textit{\TS labels (117 classes)}} \\
\quad 5 specialists & 88.21 & & 90.04 & \\
\quad 1 plain U-Net & 88.28 & +0.07 & 89.99 & $-$0.04 \\
\quad 1 ResEnc U-Net & \textbf{88.73} & +0.52 & \textbf{90.35} & +0.31$^*$ \\
\multicolumn{5}{@{}l}{\textit{\enh labels (125 classes)}} \\
\quad 6 specialists & 88.04 & & \textbf{90.66} & \\
\quad 1 plain U-Net & 87.84 & $-$0.20 & 90.26 & $-$0.40$^*$ \\
\quad 1 ResEnc U-Net & \textbf{88.48} & +0.44$^*$ & 90.60 & $-$0.06 \\
\bottomrule
\end{tabular}
\end{table}

\begin{figure*}[ht]
\centering
\includegraphics[width=0.8\linewidth]{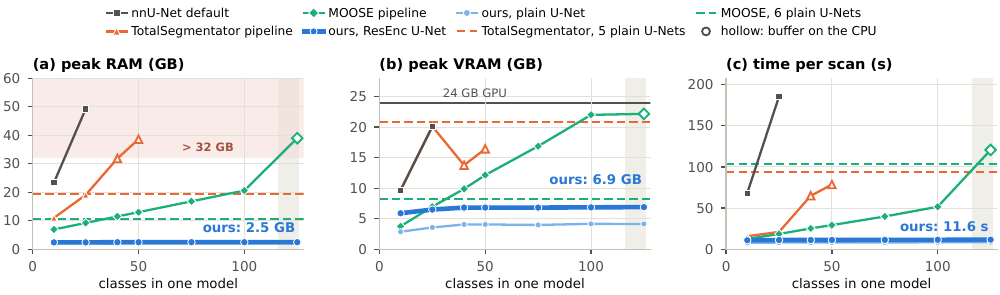}
\caption{\textbf{With our inference, time and memory barely change with the number of classes, while pipelines that hold the
logits of all classes grow until they run out of RAM.} Prediction step only, from the preprocessed image to the label map on the
scanner's grid, on a head-to-toe CT (\enh 0484), vs.\ the number of classes in one model. Every pipeline runs the same plain U-Net,
and ours also runs our ResEnc U-Net. Dashed: the toolkits' five or six networks. Gray band: the 117 and 125 classes of the two
label spaces. Shaded in (a): more RAM than a 32\,GB machine. One run per point.}
\label{fig:sweep_main}
\end{figure*}

\begin{figure}[t]
\centering
\includegraphics[width=\linewidth]{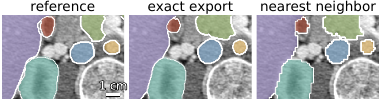}
\caption{\textbf{Nearest-neighbor export produces jagged edges.} It also lowers the \Dsc (Sec.~\ref{sec:nn}). Left to right: reference
labels, exact export and nearest-neighbor export of the same network output of our \TS-label model (1.5\,mm) on a BTCV CT with
0.59\,mm pixels. Every boundary of the nearest-neighbor export lies on the 1.5\,mm working grid.}
\label{fig:blocks}
\end{figure}

\paragraph{Datasets.}
We use two datasets. \TS v3~\cite{wasserthal2023totalsegmentator,wasserthal2026tsv3data} has 1,939 CT scans with 117
classes and an official split with 1,830 training and 109 test scans. \enh~\cite{ferrara2026enhance} has 1,683 whole-body PET/CT studies, of which we use the CT with the 125 classes of MOOSE's six networks. It has no published split, so we
redraw the 80/20 split of its dataset paper, stratified by center and diagnosis (1,346 / 337). For the export comparison in Sec.~\ref{sec:nn} we use BTCV~\cite{landman2015btcv} and AMOS~\cite{ji2022amos}.

\paragraph{Protocol.}
Our recipe was chosen by 5-fold cross-validation (CV) inside each training split, which shows consistent improvements from our
design choices (Suppl.~S8, S9). The chosen configuration is then trained on the
full training split and evaluated on the test split. Our model is a residual-encoder U-Net (ResEnc
U-Net)~\cite{isensee2024revisited} with 102\,M parameters. It is trained at 1.5\,mm with a patch size of 192$^3$, batch size 2,
one GPU and \pgps 5400 (Sec.~\ref{sec:training}; Suppl.~S9). We train the model independently for
\TS and \enh, from scratch. The specialists are the toolkits' own networks, five for \TS and six for MOOSE on \enh. Each is
\nnunet's plain convolutional U-Net (plain U-Net) with 31\,M parameters, 156\,M and 184\,M in total. We retrained them with each
toolkit's recipe on the same splits, exported their predictions exactly (linear logit interpolation, like ours) and merged them
as the toolkits do. We report the mean \Dsc over classes under two rules, FP-penalized and on present structures (defined in
Tab.~\ref{tab:onemodel}), and give paired case-bootstrap intervals for differences.

We evaluate resource scaling on a workstation with a 24\,GB GPU (RTX 4090) and 64\,GB of RAM, of which a run may use up to 55\,GB.
Inference is timed on four \enh test scans, one per site at the site's median size, and on a head-to-toe test scan
(Fig.~\ref{fig:sweep_main}). None of the four was used during development.
Fig.~\ref{fig:sweep_main} and Tab.~\ref{tab:toolkits} time only prediction, from the preprocessed image to the label
map on the scanner's grid. This is the step in which the pipelines differ in how they handle the classes. Every pipeline runs
like for like, with no \texttt{torch.compile}, no test-time augmentation and a step size of 0.5 (TotalSegmentator ships with 0.8: Suppl.~Tab.~S2). We report the median of three runs and the peak memory (1\,GB = $2^{30}$ bytes; Suppl.~S1). Sec.~\ref{sec:release} times the toolkits as released, end to end, including start-up (model loading and initialization),
reading, preprocessing and writing. TotalSegmentator and MOOSE refer to their latest releases, 2.18 and 3.2.2, unless stated otherwise.

\subsection{The cost of the class dimension}
\label{sec:sweep}
\label{sec:toolkits}
The pipelines differ in where they reduce the logits to a label map. The toolkits take the argmax at the working resolution and
upsample the label map with nearest neighbor. \nnunet's default resamples the logits of every class to the scanner's grid and
takes the argmax there. Ours computes the same label map, but streams the conversion and holds features instead of
logits (Secs.~\ref{sec:streaming}--\ref{sec:combined}). Fig.~\ref{fig:sweep_main} shows how the cost of each pipeline grows with
the number of classes. For this, every pipeline runs the same plain U-Net on a head-to-toe \enh scan (0484), with the
classification head cut to its first $C$ classes. The pipelines therefore differ only in how they handle the network's output.
We also run ours with our ResEnc U-Net, which is larger and therefore needs more VRAM and time.
Table~\ref{tab:toolkits} compares the pipelines at the full label spaces on the four test scans.
It also includes the toolkits' five or six networks (Sec.~\ref{sec:onemodel}).

\paragraph{Canonical pipelines hit a memory wall.}
Every pipeline that holds the logits of all classes at image size needs more memory with every class. On scan 0484, \nnunet's
default needs 1.7\,GB of RAM per class and runs out of RAM at 40 classes. MONAI's inference with the same plain U-Net runs out of
VRAM from 25 classes and, after its fallback to RAM, out of RAM from 40. The wall is therefore not specific to one framework. The
toolkits build on \nnunet and suffer from the same problem, which is why they split their label spaces. On this scan, VISTA3D holds its logits in 37--48\,GB of RAM at 50 and 75 classes
and runs out of RAM from 100, and VoxTell runs out of VRAM from 25 prompts, as both hold one logit map per class (Suppl.~S4).

\paragraph{Our inference removes the wall.}
The cost of our inference barely changes with the number of classes (Fig.~\ref{fig:sweep_main}). With the plain U-Net, it takes 7.8--8.5\,s on scan 0484. Its RAM stays below 3\,GB and its VRAM below 4.2\,GB. With our larger ResEnc U-Net, it takes 11--12\,s, 2.5\,GB of RAM and at most 6.9\,GB of VRAM. This VRAM has two main parts. The first is the forward pass of the network on one patch (1.1\,GB for the plain U-Net, 3.6\,GB for the ResEnc U-Net). The second is the ring, which holds a fixed number of feature slices. We stream along the image's longest axis, so the ring's size depends on the image's smallest cross-section, not on the classes (1.0--1.9\,GB on the scans of Tab.~\ref{tab:toolkits}).

\paragraph{At the full label spaces.}
On the scans of Tab.~\ref{tab:toolkits}, ours runs the plain U-Net 6.3--17$\times$ faster than the toolkits' pipelines, with or without their split. These need up to 53\,GB of RAM and 23\,GB of VRAM, where ours needs at most 3.2\,GB of RAM and 4.4\,GB of VRAM. \nnunet's default and MONAI run out of RAM. Our inference also speeds up the toolkits' own split: it runs their five or six networks 1.8--2.3$\times$ faster and with 2.5--6.3$\times$ less RAM. Running one network instead of five or six gives another 4.7--5.8$\times$.

\subsection{Exact vs.\ nearest-neighbor export}
\label{sec:nn}
To keep their pipelines tractable, the toolkits take the argmax at the working resolution and upsample the label map with nearest
neighbor. Our inference resamples the logits exactly. To measure what this shortcut costs, we export the network output of each scan in both ways and score them 
against the reference labels drawn on the scanner's grid.
On 30 BTCV scans with manual labels (13 abdominal organs, 12 for the \enh-label model), nearest neighbor costs our \TS-label
model 0.88 \Dsc points (95\,\% CI 0.70--1.06) and our \enh-label model 0.47 (0.37--0.57). It lowers the score on 30 and 28 of the
30 scans, respectively. On AMOS CTs (14 organs, 12 for the \enh-label model; scans in Suppl.~S11), it costs 0.66 (0.61--0.71) and 0.52
(0.47--0.56) points. The normalized surface Dice (NSD)~\cite{nikolov2021clinically} at a tolerance of 1\,mm drops by 1.08
and 0.88 points (BTCV, AMOS) for the \TS-label model. For the \enh-label model it changes by +0.04 and $-$0.57 (Suppl.~Tab.~S11).
Independently of any reference, nearest neighbor changes the label of 3.5\,\% (\TS-label model) and 2.8\,\% (\enh-label model)
of the foreground voxels on BTCV. It also places every boundary on the working grid, which is visible as staircases
(Fig.~\ref{fig:blocks}).

\subsection{One model instead of the split}
\label{sec:onemodel}
Without the memory wall, a label space no longer has to be split. We test on the two datasets what the split is worth
in segmentation quality and what it costs at inference.

\paragraph{Segmentation quality.} Our ResEnc U-Net matches or slightly exceeds the composed specialists on both label spaces
(Tab.~\ref{tab:onemodel}). FP-penalized, it is 0.52 \Dsc points ahead on \TS and 0.44 on \enh. On present structures, it is 0.31
points ahead on \TS and level on \enh. For the 0.44 and the 0.31, the 95\,\% bootstrap interval of the difference excludes 0. The
FP-penalized gains are larger because our model makes fewer false detections, 128 against 146 (scan, class) pairs on \TS and 387
against 467 on \enh. We did not scale the specialists' architectures or compute. The result therefore shows that the split is not
needed, but not that one model is the best possible.

\paragraph{Capacity and cost.} One model has to learn what five or six specialists learn. With our recipe, a single plain U-Net,
the specialists' own network, already matches them on \TS under both rules (Tab.~\ref{tab:onemodel}). On \enh its FP-penalized
difference is $-$0.20 points (95\,\% CI $-$0.47 to $+$0.06), and it is 0.40 points behind on present structures. Our ResEnc U-Net
is 0.3--0.6 points ahead of the plain U-Net under both rules on both label spaces, and the CV agrees (Suppl.~S8). This capacity
costs little at inference. The ResEnc U-Net takes 1.0--1.5$\times$ the time of the plain U-Net in our pipeline and is still
6.3--9.6$\times$ faster than the toolkits' five or six networks, with 2.5--6.1$\times$ less RAM (Tab.~\ref{tab:toolkits}). Its
VRAM stays below TotalSegmentator's on every scan and below MOOSE's on three of four. In the toolkits, a third to a half of the
extra time goes into the additional networks and the rest into handling the logits (Suppl.~S2).

\subsection{End to end, against the releases}
\label{sec:release}
\begin{table}[t]
\caption{\textbf{End to end, our integrations make TotalSegmentator and MOOSE 5.8--15$\times$ faster, with 3.2--7.6$\times$ less
RAM.} They run one unified ResEnc U-Net (Sec.~\ref{sec:onemodel}), larger than each of the five or six plain U-Nets it replaces, with
our inference in an optimized pipeline (Suppl.~S5). With the same model, \nnunet 2.8.1 runs out of RAM, and \nnunet with our
inference takes 10--20\,s. Time from launching the command to the label map on disk, with each tool's shipped settings, in s; memory
in GB. Ranges over the scans of Tab.~\ref{tab:toolkits}. Folder: time per scan for 14 \enh scans in one call. TotalSegmentator runs
at its step 0.8, all others at 0.5.}
\label{tab:release}
\centering
\footnotesize
\setlength{\tabcolsep}{3pt}
\begin{tabular}{@{}llccccc@{}}
\toprule
 & & \multicolumn{3}{c}{GPU, one scan} & folder & CPU \\
\cmidrule(lr){3-5}\cmidrule(lr){6-6}\cmidrule(l){7-7}
\multicolumn{2}{@{}l}{pipeline (networks)} & time & RAM & VRAM & time & time \\
\midrule
\multicolumn{7}{@{}l}{\textbf{\TS labels (117 classes)}} \\
 & 2.18 (5) & 69--162 & 13--32 & 9.6--21 & 77 & 510--1630 \\
\rowcolor{oursrow} \cellcolor{white}\multirow{-2}{*}{\shortstack[l]{TS}} & ours & 5.8--11 & 2.5--6.0 & 6.1--7.2 & 2.5 & 120--325 \\
\addlinespace[3pt]
 & 2.8.1 & \multicolumn{5}{c}{\textit{out of RAM}} \\
\rowcolor{oursrow} \cellcolor{white}\multirow{-2}{*}{\nnunet} & ours & 11--20 & 2.6--6.4 & 6.1--7.3 & 4.5 & 343--811 \\
\midrule
\multicolumn{7}{@{}l}{\textbf{\enh labels (125 classes, MOOSE)}} \\
 & 3.2.2 (6) & 67--115 & 12--25 & 4.5--8.5 & 68 & 1773--3546 \\
\rowcolor{oursrow} \cellcolor{white}\multirow{-2}{*}{MOOSE} & ours & 9.7--18 & 2.5--7.8 & 6.1--7.3 & 4.7 & 314--741 \\
\addlinespace[3pt]
 & 2.8.1 & \multicolumn{5}{c}{\textit{out of RAM}} \\
\rowcolor{oursrow} \cellcolor{white}\multirow{-2}{*}{\nnunet} & ours & 10--19 & 2.4--5.8 & 6.1--7.3 & 4.5 & 346--812 \\
\bottomrule
\end{tabular}
\end{table}

We integrated our inference into \nnunet and into both toolkits. In the toolkits, one unified ResEnc U-Net model per label space replaces the
five or six plain U-Net networks, and exact resampling replaces the nearest-neighbor upsampling of labels. In \nnunet, every model gets exact
streaming inference as a drop-in, whether it is existing or new.
Table~\ref{tab:release} compares the integrations with the released versions end to end, as users run them.

\paragraph{Faster end to end.} On a single scan (Tab.~\ref{tab:release}), our integration into \TS is 12--15$\times$ faster
than its release and our integration into MOOSE 5.8--7.6$\times$. These factors have two sources. One model instead of five or six, run with our inference,
makes prediction 6.3--9.6$\times$ faster (Tab.~\ref{tab:toolkits}), and both tools get optimized pipelines (Suppl.~S5).
TotalSegmentator's unreleased development branch narrows the gap, but ours is still 3.5--4.7$\times$ faster than it (Suppl.~S6). On a folder of 14 scans,
the factors grow to 31 and 15$\times$, and a scan takes 2.5\,s in \TS and 4.7\,s in MOOSE. On the CPU, ours is 3.7--7.3$\times$
faster than \TS and 4.7--5.7$\times$ faster than MOOSE. In \nnunet, our streaming inference needs 10--20\,s per scan, whereas
\nnunet 2.8.1 runs out of RAM on every scan.

\paragraph{Less memory.} On the GPU, ours needs 3.2--7.6$\times$ less RAM than TotalSegmentator and MOOSE. It needs 1.6--2.9$\times$ less VRAM than \TS and about as much as MOOSE (0.6--1.3$\times$ ours).

\section{Discussion and conclusion}
\label{sec:conclusion}

\paragraph{Generality and adoption.}
Streaming needs only sliding-window inference. Holding features instead of logits (Proposition~1) needs a pointwise affine head
and only weighted averages between the network and the argmax.
These conditions are not specific to medical images. Any dense predictor that blends overlapping windows and ends in a pointwise
affine head meets them, for example in volumetric microscopy or for voxelized 3D scenes. We have not tested such models. The change is local and needs no retraining, so it could be integrated into MONAI and
other frameworks. Our predictor for \nnunet applies to every model trained with that framework, and in TotalSegmentator and MOOSE one
model replaces the five and six networks. Splitting a label space thereby becomes a modeling choice.
Streaming also allows higher working resolutions and larger images than before.

\paragraph{One model, all labels.}
One observation surprised us. A single model sees every label and could exploit the relations between structures that the split
separates. Yet it only matches specialists that each learn a subset, and it needs a large patch and a long schedule (Suppl.~S8). Whether this reflects optimization, capacity or rare classes is open. Without the memory wall, this can now be studied.

\paragraph{Limitations.}
Every CV fold and every test-split model is one training run. The specialists were trained with their toolkits' recipes. We did
not scale their architectures or compute, train them with \pgps or extend their schedules. Fully tuned specialists may therefore
still segment better than one model and could be preferred. Our
implementation is based on \nnunet, and we timed it on one machine. Our claims about consumer hardware concern inference, not
training (Sec.~\ref{sec:training}).

\paragraph{Conclusion.}
The memory wall of many-class 3D segmentation lies in what the pipeline materializes. Two observations remove it. First, every
step after the network is local along one axis, so the logits can be streamed instead of being held for the whole volume. Second, the head is affine and commutes with blending, mirroring and resampling, so the stream can carry the network's features instead of the logits. Together, they keep the class count from ever multiplying the volume, and the label map stays exact.
This has two consequences. Existing models get exact many-class inference without retraining. Unified models become deployable,
and one model per label space runs on consumer hardware, faster than the split pipelines and matching or slightly exceeding their specialists. 

\section*{LLM Usage Statement}
We used Claude Opus 5.5 and Claude Fable 5.2 (Anthropic) as coding and writing assistants. They contributed to the training, inference and benchmark code, to
running and analyzing experiments, to figures and tables, and to editing the text. We used GPT-6.1 Sol and GPT-6 Astra
(OpenAI) for figures, and GPT-6.1 Sol also to review the manuscript and to edit the text. The authors directed the work, reviewed the
code and text, verified all reported results against the raw outputs and take full responsibility for the content.

\section*{Acknowledgments}
This work was supported by the Helmholtz Association under the Helmholtz Foundation Model Initiative (HFMI), project
``The Human Radiome Project'' (THRP).
This work was supported by the NCT Heidelberg Image Data Analysis Infrastructure (IDAI).
Maximilian Rokuss was supported by the Google PhD Fellowship Program.
Stefan Michael Happold acknowledges funding by the Deutsche Forschungsgemeinschaft (DFG, German Research Foundation) --
515279324/SPP 2177.
Stefan Michael Happold acknowledges funding from the Free State of Bavaria (StMGP) as part of the project
`KI-gest\"utzte multi-modale Diagnostik und stratifizierte Therapie f\"ur Endometriose (EndoKI)'.

{
    \small
    \bibliographystyle{ieeenat_fullname}
    \bibliography{references}
}

\clearpage
\onecolumn
\setcounter{section}{0}\setcounter{figure}{0}\setcounter{table}{0}
\renewcommand{\theHsection}{S\arabic{section}}
\renewcommand{\theHtable}{S\arabic{table}}
\renewcommand{\theHfigure}{S\arabic{figure}}
\begin{center}{\Large\bf Supplementary Material}\end{center}
\vspace{12pt}
\renewcommand{\thetable}{S\arabic{table}}
\renewcommand{\thesection}{S\arabic{section}}
\renewcommand{\thefigure}{S\arabic{figure}}

\section{How inference was measured}
\label{sec:supp_measure}
Tab.~1 and Fig.~3 of the main paper and the class sweeps here time prediction: the step from the preprocessed
image, held in memory, to the integer label map on the scanner's grid, also in memory. The comparison targets the conceptual
differences between the approaches, so everything around this step is either left out or the same for every pipeline. TotalSegmentator and MOOSE refer to their latest releases, 2.18 and 3.2.2, unless stated otherwise.

\paragraph{Timed.} All of a pipeline's networks, with their sliding-window forward passes, blending, class head, resampling and
argmax, in the order and on the device the pipeline uses. For the toolkits this includes their conversion of each network's
logits to a label map and their steps on the label maps after the networks: TotalSegmentator resamples them to the scanner's
grid and undoes its reorientation, and MOOSE resamples them. For ours it is the single fused pass and the reversal of cropping
and reorientation. Our predictor keeps the preprocessed input, the ring and the head on the GPU, and the fused kernel runs there
as well.

\paragraph{Not timed.} Model loading, reading, preprocessing including the resampling of the input, and writing. The Gaussian
tile weighting, which is setup (0.2--1.4\,s per scan and pipeline). One forward pass on an all-zero patch before each network's
first prediction, in which cuDNN searches its convolution algorithms. Compilation, since every pipeline runs eager.
TotalSegmentator's merge of its five part label maps, which is interleaved with reading the part files. Leaving it out favors
TotalSegmentator.

\paragraph{Conversion in the prediction process.} \nnunet 2.8.1 and the toolkits built on it send each network's full logit
tensor to a worker process, which converts it to a label map. We call the same conversion functions in the prediction process
instead. The computation is unchanged, and the transfer between processes disappears. On the head-to-toe scan of Fig.~3,
TotalSegmentator as shipped spends about 37\,s of its 151\,s on this transfer (14\,s pickling in the main process, 23\,s
receiving in the export workers).

\paragraph{Settings.} The same for every pipeline: eager, cuDNN's benchmark mode on, step 0.5 unless stated, no test-time
augmentation, mixed precision as each pipeline ships it, one fold per network. \nnunet 2.8.1 and the toolkits switch the
benchmark mode on themselves. Our predictor runs with it off by default, and on TotalSegmentator's network it makes our forward passes
1.6\,\% faster.

\paragraph{Same label maps.} We checked that the timed code produces each pipeline's own label map. The label maps are identical
for MOOSE, its pipeline with one network, \nnunet 2.8.1 on MOOSE's network, the TotalSegmentator pipeline with one network
and the TotalSegmentator development branch. For TotalSegmentator 2.18, \nnunet 2.8.1 on TotalSegmentator's network and ours, at most 0.0012\,\% of
the voxels differ, because cuDNN picks different algorithms in different processes. VISTA3D's label maps differ in
0.0003\,\% of the voxels.

\paragraph{Memory.} RAM is the peak proportional set size of the prediction process while the timed steps run, sampled at 5\,Hz.
Idle worker processes are not counted. VRAM is the larger of the process's usage reported by nvidia-smi (sampled at 4\,Hz,
including the CUDA context) and PyTorch's peak reserved memory during the timed steps. 1\,GB = $2^{30}$ bytes.

\paragraph{Protocol.} Every run starts a fresh process. Tab.~1: three repeats per scan, interleaved across pipelines, median
time and maximum memory. Class sweeps: one run per point. A run may use up to 55\,GB of RAM with swap off, and a run that needs
more is reported as out of RAM. Workstation with an RTX 4090 (24\,GB), a Ryzen 9 7950X (16 cores) and 64\,GB of RAM.

\section{Tab.~1 per scan}
Table~\ref{tab:toolkits_scans} gives the numbers behind the ranges of Tab.~1 of the main paper for each scan.
\begin{table}[h]
\caption{\textbf{Tab.~1 of the main paper per scan:} prediction on one \enh test scan from each of its four
sites (the site's median size, field of view and slice thickness above). Time in s (median of three runs), peak RAM and peak VRAM
in GB, every pipeline eager at step 0.5 (Sec.~\ref{sec:supp_measure}). Plain U-Net: \nnunet's plain convolutional U-Net, the
network of each toolkit specialist, here for all classes. Ours with the toolkit's plain U-Nets: each network through our inference, the label maps merged as in the toolkit. $^\dagger$The logit buffer does not fit in VRAM, and the pipeline restarts
with it in RAM, here repeating every window. $^\ast$MONAI 1.6.1 with its automatic fallback (setup with Fig.~\ref{fig:sweep}): the
logit buffer (24--67\,GB) does not fit in VRAM, MONAI moves it to RAM, and then resampling the logits to the scanner's grid
(0371, 1046) or adding up the windows in RAM (1412, 1600) needs more; with MONAI's defaults the buffer stays on the GPU and runs
out of VRAM. Out of RAM: more than the 55\,GB available to each run. Bold: the lowest value per
label set.}
\label{tab:toolkits_scans}
\centering
\footnotesize
\setlength{\tabcolsep}{2.4pt}
\newcommand{\fail}[2]{\multicolumn{#1}{c}{\textit{#2}}}
\newcommand{\fb}{$^\dagger$}
\resizebox{\textwidth}{!}{%
\begin{tabular}{l*{12}{N{0.86cm}}}
\toprule
 & \multicolumn{3}{c}{AutoPET (0371)} & \multicolumn{3}{c}{Leipzig (1046)} & \multicolumn{3}{c}{Careggi (1412)} & \multicolumn{3}{c}{UC Davis (1600)} \\
 & \multicolumn{3}{c}{44$\times$44$\times$98\,cm, 2.5\,mm} & \multicolumn{3}{c}{50$\times$50$\times$98\,cm, 3\,mm} & \multicolumn{3}{c}{60$\times$60$\times$86\,cm, 5\,mm} & \multicolumn{3}{c}{50$\times$50$\times$194\,cm, 1\,mm} \\
\cmidrule(lr){2-4}\cmidrule(lr){5-7}\cmidrule(lr){8-10}\cmidrule(lr){11-13}
pipeline & time & RAM & VRAM & time & RAM & VRAM & time & RAM & VRAM & time & RAM & VRAM \\
\midrule
\multicolumn{13}{l}{\textbf{\TS label set (117 classes)}} \\
\TS 2.18, 5 plain U-Nets & 33.1 & 8.2 & 9.6 & 46.6 & 10.3 & 11.3 & 55.6 & 12.4 & 14.3 & 95.0 & 20.6 & 20.7 \\
\TS dev.\ branch (Oct.\ 2026), 5 plain U-Nets & 19.8 & 8.1 & 9.8 & 29.7 & 9.0 & 12.0 & 34.2 & 9.8 & 14.6 & 60.5 & 18.5 & 21.4 \\
\rowcolor{oursrow} \textbf{ours}, the 5 plain U-Nets of \TS 2.18 & 16.5 & 2.2 & \textbf{2.5} & 25.5 & 2.9 & \textbf{2.7} & 29.3 & 2.0 & \textbf{3.1} & 52.0 & 5.1 & \textbf{3.3} \\
\cmidrule(l){1-13}
\TS pipeline, 1 plain U-Net & 49.9\fb & 33.3 & 15.3 & 72.3\fb & 42.9 & 19.1 & 103.5\fb & 53.4 & 23.4 & \fail{3}{out of RAM} \\
\nnunet 2.8.1, 1 plain U-Net & \fail{12}{out of RAM on all four scans} \\
MONAI, 1 plain U-Net & \fail{12}{out of RAM on all four scans$^\ast$} \\
\rowcolor{oursrow} \textbf{ours}, 1 plain U-Net & \textbf{3.4} & \textbf{1.7} & 2.7 & \textbf{5.3} & \textbf{1.8} & 2.9 & \textbf{6.0} & \textbf{1.8} & 3.6 & \textbf{11.0} & \textbf{3.2} & 3.8 \\
\cmidrule(l){1-13}
\rowcolor{oursrow} \textbf{ours}, 1 ResEnc U-Net & 5.3 & 2.1 & 6.1 & \textbf{5.3} & 2.2 & 6.5 & 7.7 & 2.3 & 7.3 & 11.9 & 3.4 & 7.2 \\
\midrule
\multicolumn{13}{l}{\textbf{\MOOSE label set (125 classes)}} \\
MOOSE, 6 plain U-Nets & 45.5 & 7.5 & 6.8 & 51.2 & 8.5 & 8.3 & 69.8 & 5.6 & 4.6 & 105.7 & 14.3 & 8.5 \\
\rowcolor{oursrow} \textbf{ours}, the 6 plain U-Nets of MOOSE & 22.3 & 2.4 & \textbf{3.0} & 22.3 & 2.4 & \textbf{3.2} & 39.7 & 2.1 & \textbf{3.3} & 49.5 & 5.7 & \textbf{4.1} \\
\cmidrule(l){1-13}
\MOOSE pipeline, 1 plain U-Net & 25.0 & 17.3 & 20.4 & 59.3\fb & 35.1 & 21.1 & 46.0 & 10.4 & 11.6 & 121.4\fb & 41.6 & 22.1 \\
\nnunet 2.8.1, 1 plain U-Net & \fail{12}{out of RAM on all four scans} \\
MONAI, 1 plain U-Net & \fail{12}{out of RAM on all four scans$^\ast$} \\
\rowcolor{oursrow} \textbf{ours}, 1 plain U-Net & \textbf{4.0} & \textbf{1.9} & 3.3 & \textbf{4.0} & \textbf{1.7} & 3.6 & \textbf{6.8} & \textbf{1.8} & 3.7 & \textbf{8.9} & \textbf{2.0} & 4.4 \\
\cmidrule(l){1-13}
\rowcolor{oursrow} \textbf{ours}, 1 ResEnc U-Net & 5.3 & 2.1 & 6.1 & 5.3 & 2.2 & 6.6 & 7.7 & 2.3 & 7.3 & 11.9 & 2.5 & 7.2 \\
\bottomrule
\end{tabular}%
}
\end{table}

Table~\ref{tab:step08} repeats it at TotalSegmentator's step 0.8 (the main paper runs every pipeline at 0.5).
\begin{table}[h]
\caption{\textbf{Tab.~1 at TotalSegmentator's step 0.8,} \TS label set, same scans and protocol as Tab.~\ref{tab:toolkits_scans}:
time in s (median of three runs), peak RAM and VRAM in GB. On Leipzig and UC Davis ours is faster with the ResEnc U-Net than with
the plain U-Net: at this step its 192$^3$ windows cover these scans with 48 and 58\,\% of the voxels of the plain U-Net's 128$^3$
windows (Leipzig: 16 instead of 112 windows; at step 0.5, 54 and 250 windows take the same time).}
\label{tab:step08}
\centering
\footnotesize
\setlength{\tabcolsep}{2.4pt}
\resizebox{\textwidth}{!}{%
\begin{tabular}{l*{12}{N{0.86cm}}}
\toprule
 & \multicolumn{3}{c}{AutoPET (0371)} & \multicolumn{3}{c}{Leipzig (1046)} & \multicolumn{3}{c}{Careggi (1412)} & \multicolumn{3}{c}{UC Davis (1600)} \\
\cmidrule(lr){2-4}\cmidrule(lr){5-7}\cmidrule(lr){8-10}\cmidrule(lr){11-13}
pipeline & time & RAM & VRAM & time & RAM & VRAM & time & RAM & VRAM & time & RAM & VRAM \\
\midrule
\TS 2.18, 5 plain U-Nets & 23.4 & 8.2 & 9.6 & 32.9 & 10.3 & 11.8 & 36.4 & 12.4 & 13.9 & 65.8 & 20.6 & 20.7 \\
\TS dev.\ branch (Oct.\ 2026), 5 plain U-Nets & 10.0 & 7.9 & 9.7 & 15.6 & 9.0 & 11.6 & 14.9 & 9.8 & 14.6 & 31.1 & 18.5 & 21.4 \\
\rowcolor{oursrow} \textbf{ours}, the 5 plain U-Nets of \TS 2.18 & 6.8 & 2.3 & 2.5 & 11.6 & 2.1 & 2.7 & 9.9 & 2.0 & 3.1 & 22.6 & 5.3 & 3.3 \\
\rowcolor{oursrow} \textbf{ours}, 1 plain U-Net & 1.5 & 1.7 & 2.7 & 2.5 & 1.7 & 3.0 & 2.2 & 1.8 & 3.6 & 5.1 & 2.0 & 3.8 \\
\rowcolor{oursrow} \textbf{ours}, 1 ResEnc U-Net & 1.7 & 2.1 & 6.2 & 1.8 & 2.2 & 6.5 & 3.6 & 2.3 & 7.3 & 4.3 & 2.5 & 7.2 \\
\bottomrule
\end{tabular}%
}
\end{table}

Table~\ref{tab:breakdown} splits the times of Table~\ref{tab:toolkits_scans} into the networks' forward passes and the handling of
the logits. Against ours on the same plain U-Net, the additional networks account for 42--47\,\% of the extra time of \TS
2.18 and 36--51\,\% of that of MOOSE, and the logits for the rest; the development branch handles its logits faster, so for it the
networks account for 76--80\,\%.
\begin{table}[h]
\caption{\textbf{Where the toolkits' time goes:} prediction of Tab.~\ref{tab:toolkits_scans} split into the
networks' forward passes (column net.) and everything that handles the logits (column logits), in s, each the median of three runs. For the
toolkits, each network's logits are blended into a buffer with one channel per class, moved to the host and converted to a label
map, which is then brought to the scanner's grid; ours does all of this within its single pass.}
\label{tab:breakdown}
\centering
\footnotesize
\setlength{\tabcolsep}{2.4pt}
\resizebox{\textwidth}{!}{%
\begin{tabular}{l*{12}{N{0.86cm}}}
\toprule
 & \multicolumn{3}{c}{AutoPET (0371)} & \multicolumn{3}{c}{Leipzig (1046)} & \multicolumn{3}{c}{Careggi (1412)} & \multicolumn{3}{c}{UC Davis (1600)} \\
\cmidrule(lr){2-4}\cmidrule(lr){5-7}\cmidrule(lr){8-10}\cmidrule(lr){11-13}
pipeline & net. & logits & total & net. & logits & total & net. & logits & total & net. & logits & total \\
\midrule
\multicolumn{13}{l}{\textbf{\TS label set (117 classes)}} \\
\TS 2.18, 5 plain U-Nets & 15.4 & 17.6 & 33.1 & 24.1 & 22.5 & 46.6 & 27.8 & 27.8 & 55.6 & 48.3 & 46.9 & 95.0 \\
\TS dev.\ branch (Oct.\ 2026), 5 plain U-Nets & 15.5 & 4.4 & 19.8 & 24.2 & 5.5 & 29.7 & 27.8 & 6.4 & 34.2 & 48.4 & 12.1 & 60.5 \\
\rowcolor{oursrow} \textbf{ours}, 1 plain U-Net & 3.0 & 0.4 & 3.4 & 4.8 & 0.5 & 5.3 & 5.5 & 0.5 & 6.0 & 9.5 & 1.4 & 11.0 \\
\rowcolor{oursrow} \textbf{ours}, 1 ResEnc U-Net & 4.9 & 0.4 & 5.3 & 4.9 & 0.5 & 5.3 & 7.2 & 0.5 & 7.7 & 10.5 & 1.3 & 11.9 \\
\midrule
\multicolumn{13}{l}{\textbf{\MOOSE label set (125 classes)}} \\
MOOSE, 6 plain U-Nets & 24.7 & 20.7 & 45.5 & 24.8 & 26.4 & 51.2 & 28.4 & 41.3 & 69.8 & 49.5 & 56.1 & 105.7 \\
\rowcolor{oursrow} \textbf{ours}, 1 plain U-Net & 3.4 & 0.6 & 4.0 & 3.4 & 0.6 & 4.0 & 6.1 & 0.7 & 6.8 & 7.5 & 1.4 & 8.9 \\
\rowcolor{oursrow} \textbf{ours}, 1 ResEnc U-Net & 4.9 & 0.4 & 5.3 & 4.9 & 0.5 & 5.3 & 7.2 & 0.5 & 7.7 & 10.5 & 1.4 & 11.9 \\
\bottomrule
\end{tabular}%
}
\end{table}

\section{Where the class count enters the cost}
Table~\ref{tab:cost} expands Secs.~3.1 and~4 of the main paper stage by stage.
\begin{table*}[t]
\caption{\textbf{Where the class count enters the cost} of an \nnunet pipeline at 117--125 classes, stage by stage (training
measurements: 192$^3$, batch 2, one L40S; inference: one RTX 4090 workstation). The network itself carries $C$ only in its
final $1{\times}1{\times}1$ convolutions. Training memory and time are PyTorch's peak allocation and the duration of the training
step (250 iterations per epoch), as in Fig.~\ref{fig:traincost}: ours takes 135\,s per epoch at 24 classes and 178\,s at 117.
Whole epochs, with their 50 validation iterations, take 145 and 189\,s, and nvidia-smi's peak over an epoch is 32\,GB at 117
classes. Inference row: prediction (Sec.~\ref{sec:supp_measure}) on the two scans of Fig.~\ref{fig:sweep}, ours
with the ResEnc U-Net; deployment row: end to end (Sec.~\ref{sec:supp_release}).}
\label{tab:cost}
\centering
\small
\setlength{\tabcolsep}{4pt}
\begin{tabular}{@{}L{2.0cm}L{3.6cm}L{5.4cm}L{5.4cm}@{}}
\toprule
pipeline stage & what holds the class dimension & \nnunet 2.8.1, 117--125 classes & ours \\
\midrule
sampling data & per-class foreground locations, one file per case & 23\,/\,41\,GB (\TS\,/\,\enh, all cases) & one compact store: 0.52\,/\,0.97\,GB (44\,/\,42$\times$ smaller) \\
augmentation & label map resampled as $C$ one-hot channels (CPU) & CPU per sample 0.8\,s at 1 class, 4.6\,s at 117 & nearest-neighbor label-map resampling: 0.4--0.6\,s at any class count; similar \Dsc (11 specialist pairs, Table~\ref{tab:paired}) \\
loss & softmax, one-hot and soft-Dice terms at every deep-supervision scale & +77\,ms and +6.3\,GB per training step at 117 classes (vs.\ ours) & whole loss compiled into fused kernels; training step at 117 classes: 30\,GB peak on a 48\,GB GPU \\
CE backward & int32 indexing: batch$\times C\times$voxels $\geq 2^{31}$ & illegal memory access from 3 samples per GPU at 192$^3$ & chunked over the batch; inert below the limit \\
validation step & one-hot prediction of the patch, $(B, C, \text{patch})$ & out of a 48\,GB GPU from 96 classes & $C\times C$ confusion matrix from the argmax (int64: 109\,/\,124\,KB for \TS\,/\,\enh) \\
final validation & $C$-channel probabilities per case at native resolution & killed for lack of RAM at 10--31 classes on native geometry & streaming predictor \\
inference & $C$-channel logits for the image, an fp32 copy, resampled logits on the native grid & 29.7\,+\,46.5\,GB (fp32) for a median whole-body CT at 125 classes; out of RAM from 75 classes (40 on a head-to-toe scan) & feature ring and slice-wise head: 2.1--2.3\,GB RAM (2.4--2.5\,GB on the head-to-toe scan) at any class count in prediction; same labels up to floating-point rounding \\
deployment & 5--6 models, nearest-neighbor label upsampling & \TS 2.18: 13--32\,GB RAM, 69--162\,s per scan & one model, exact export: 12--15$\times$ faster, 4.8--7.6$\times$ less RAM \\
training time & many-class models need long schedules & dense 5000 epochs: 5 dense-1000 units ($\sim$240 L40S GPU-h) & \pgps 5400: 1.64 units ($\sim$80 GPU-h); in 5-fold CV within 0.1 \Dsc points of dense 5000 or above \\
\bottomrule
\end{tabular}
\end{table*}

\section{The class sweep on both scans}
\begin{figure}[h]
\centering
\includegraphics[width=\linewidth]{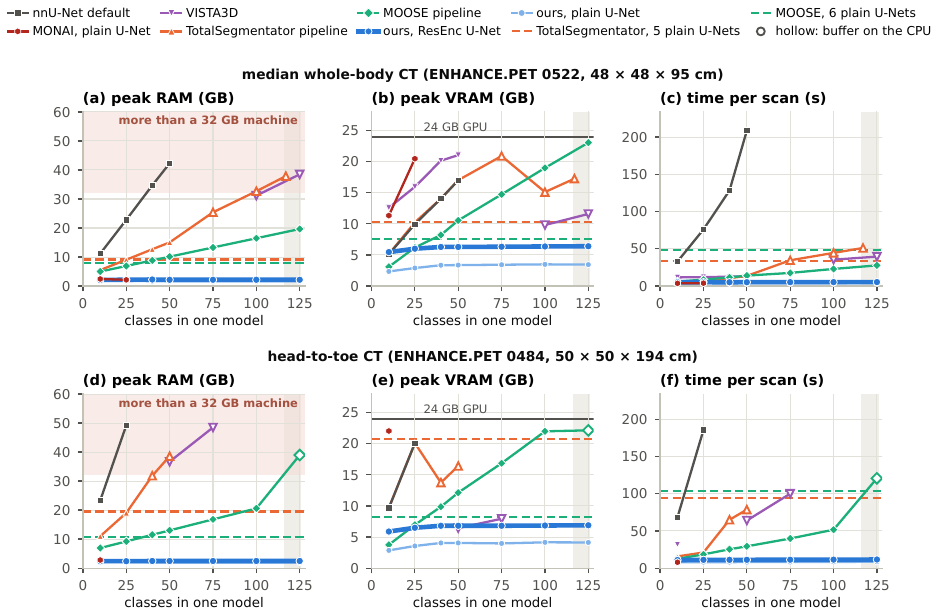}
\caption{\textbf{The class sweep on two scans:} peak RAM, peak VRAM and time of prediction vs.\ the classes in one
model, for the median-size scan (top) and the head-to-toe scan of Fig.~3 (bottom), one run per point. Lines as in Fig.~3, plus
VISTA3D (its own model) and MONAI's inference with the same plain U-Net. A line ends where the pipeline ran out of RAM; MONAI's
ends where its resampling of the logits runs out of VRAM. VISTA3D's gaps are crashes, out of VRAM in its
post-processing. Hollow: the logits were moved to RAM. In VRAM, \nnunet's default lies on the TotalSegmentator pipeline where both run (the
same \nnunet 2.8.1 sliding window on the GPU). On the head-to-toe scan, MONAI's point at 10 classes covers VISTA3D's
(22\,GB of VRAM). Ours with the ResEnc U-Net at all 125 classes: 2.2 / 2.5\,GB of RAM, 6.4 /
6.9\,GB of VRAM, 5.3 / 11.6\,s (top / bottom).}
\label{fig:sweep}
\end{figure}

Figure~\ref{fig:sweep} repeats the class sweep of Fig.~3 of the main paper on the median-size whole-body CT of \enh (0522,
48\,cm $\times$ 48\,cm $\times$ 95\,cm) in the same protocol (Sec.~\ref{sec:supp_measure}) and adds VISTA3D. On the smaller scan the
walls move to higher class counts: \nnunet's default runs out of RAM at 75 classes (head-to-toe scan: 40), and the TotalSegmentator
pipeline moves its logits to RAM from 75 classes (head-to-toe scan: 40), where it repeats its windows. The MOOSE pipeline finishes
all 125 classes on the GPU, at 23\,GB of VRAM. Ours stays flat on both scans.

\paragraph{VISTA3D and VoxTell.} Both also hold one logit map per class at image size. VISTA3D (its own model in automatic mode,
prompted with the first $N$ of its classes; protocol and label-map check in Sec.~\ref{sec:supp_measure}) runs on the GPU on the
median-size scan up to 50 classes, with its VRAM growing from 12.6 to 21.1\,GB; at 75 classes its post-processing runs out of VRAM,
and at 100 and 125 classes it holds its logits in RAM (31 and 38\,GB). On the head-to-toe scan it runs 10 classes on the GPU
(22.0\,GB of VRAM), crashes at 25 and 40 classes in its post-processing, holds its logits in RAM at 50 and 75 (37 and 48\,GB) and
runs out of RAM from 100; with the 117 classes of its automatic mode it needs 27--41\,GB of RAM on three scans of Tab.~\ref{tab:toolkits_scans} and
runs out of RAM on the fourth (UC Davis). VoxTell (release 0.1.2, one text prompt per class) keeps one logit map per prompt on the
scanner's grid in VRAM. We measured it only end to end, in an earlier measurement that ran its released command (step 0.5) on whole
scans, so it is not in Fig.~\ref{fig:sweep}: it ran 10, 25 and 40 prompts on the median-size scan with 12, 19 and 22\,GB of VRAM
and ran out of VRAM from 50 prompts, ran only 10 prompts on the head-to-toe scan (17\,GB), and ran out of VRAM on all four scans of
Tab.~\ref{tab:toolkits_scans} with the 117 \TS classes as prompts.

\paragraph{MONAI.} We also ran the same plain U-Net, with the same weights and preprocessing, through MONAI's own inference:
MONAI 1.6.1's sliding-window inferer with an automatic fallback (SlidingWindowInfererAdapt, which VISTA3D uses; overlap 0.5 and a
Gaussian with $\sigma$ = 1/8 of the window, as in \nnunet), then MONAI's inversion of the resampling on the logits (what its
Invertd transform runs, in single precision instead of its double-precision default) and the argmax. On the median-size scan it
runs 10 and 25 classes and on the head-to-toe scan 10, about as fast as ours (3.6--7.9 vs.\ 3.6--7.8\,s) and with 2.1--2.8\,GB
of RAM (ours 1.8--2.1), but its VRAM grows with the classes, to 20 and 22\,GB (ours 2.3--2.9). With more classes the logits still fit on the GPU, and resampling them there runs out of VRAM: the fallback
covers only the logit buffer. Once the buffer itself no longer fits, MONAI moves it to RAM, and the run runs out of RAM (from 100
and 40 classes). With MONAI's defaults (sliding\_window\_inference with the buffer on the GPU, double precision), the head-to-toe
scan runs out of VRAM already at 10 classes.

\paragraph{Everything on the GPU.} The canonical pipeline can also keep every step on the GPU: the logit buffer, the resampling
and the argmax. We ran it that way with \nnunet's own sliding window, resampling function and argmax, the same plain U-Net and the
protocol of Sec.~\ref{sec:supp_measure}. Its label map is identical to that of \nnunet's default, which converts the logits on
the CPU. A second variant resamples in half precision, which is the least VRAM this pipeline needs without restructuring it.
Where the pipeline fits on the GPU, it is about as fast as ours. At 10 classes the times agree within 2\,\% (3.6\,s on the
median-size scan, 7.8--8.0\,s on the head-to-toe scan), and at 25 to 50 classes ours takes 1--11\,\% less time. Its VRAM, however,
grows with the classes, up to 22\,GB, while ours needs 2.3--4.2\,GB at every class count. The pipeline runs out of VRAM from 40 classes on
the median-size scan and from 25 on the head-to-toe scan, and with half-precision resampling from 75 and 40. At 125 classes the
two variants would need 91 and 38\,GB of VRAM on the median-size scan and 195 and 81\,GB on the head-to-toe scan, computed from
the tensor sizes. Streaming is therefore only slightly faster than inference with everything on the GPU. Its benefit is the VRAM:
it makes exact and fast inference possible on consumer GPUs at any class count.

\section{The toolkits end to end}
\label{sec:supp_release}
Sec.~5.4 of the main paper compares the toolkits as released with the same toolkits running our inference and our model;
Table~\ref{tab:release_scans} gives its numbers per scan.

\paragraph{Pipelines.} The baselines are \TS 2.18.0, the \TS development branch (Sec.~\ref{sec:supp_tsv3}), MOOSE 3.2.2 (package moosez) and \nnunet 2.8.1,
each called through its own command with its own settings: step 0.8 in TotalSegmentator, 0.5 in MOOSE and \nnunet, no test-time
augmentation (every network was trained without mirroring), compilation and memory allocator as shipped. Ours are the same two
toolkits with our inference and one model per label set (the models of Tab.~2 of the main paper) in place of their networks, and
our command line for \nnunet with either model. They keep the toolkits' step sizes, compile the network when given more than
four scans (with a warm compile cache, as after a first use) and run eagerly on a single scan. On the same scan and weights, the
label maps of both integrations equal those of our predictor called directly (no voxel differs). \nnunet 2.8.1 runs our model with
its own predictor.

\paragraph{Timing.} Time runs from launching the process to its exit: interpreter start-up, imports, model loading, reading,
preprocessing, prediction, conversion, writing and teardown. One scan: three runs per scan, interleaved across pipelines, median
time. Folder: 14 further \enh test scans in one process (TotalSegmentator, which has no folder mode: one call of its Python interface
per scan), two runs, median time divided by 14. CPU: one run per scan with the GPU
hidden. The toolkits' CPU rows come from an earlier run with the same code on the same machine; repeating the TotalSegmentator development branch
on the Careggi scan took 717.9\,s, against 718.8\,s then. On the CPU, \nnunet 2.8.1 reached the 55\,GB cap on two scans and was terminated
there; we count these runs as out of RAM.

\paragraph{Memory.} RAM is the peak proportional set size of the process and its children, sampled at 5\,Hz; VRAM is the peak sum
of their usage as reported by nvidia-smi. Both are the largest over the runs, because on the folder the peak depends on which
large scans meet in the pipeline, and because nvidia-smi samples at 4\,Hz and can miss short peaks (\TS 2.18 on UC Davis: 10.4\,GB
in one run, against 20.7\,GB in PyTorch's own count); the largest of three runs agrees with PyTorch's reserved memory plus the
CUDA context within 0.5\,GB. In the folder runs, MOOSE and its integration run as a subprocess of the measuring process; we
time that subprocess from launch to exit and subtract the measuring process's own share, its idle CUDA context (0.45\,GB of VRAM,
about 0.6\,GB of RAM). On the folder, \TS 2.18, the development branch and ours need 30, 16 and 6.0\,GB of RAM and 21, 21 and 6.7\,GB of VRAM;
MOOSE and ours 27 and 7.8\,GB of RAM and 8.5 and 6.7\,GB of VRAM. No run was disturbed by other work on the GPU (no other process
above 20\,\% of the GPU's compute in three or more samples).

\paragraph{Where the time goes.} On one scan on the GPU, \TS 2.18 spends 17--37\,s reading and preparing the input and 37--90\,s
on the logits and label maps after its networks, MOOSE 16--30\,s after its networks; ours spends under 1\,s after its
network. Ours reads and preprocesses the scan in a background thread while the model loads. On three scans
this hides most of it (the prediction waits at most 0.3\,s); on the largest, UC Davis, it
waits 1.2--2.3\,s, mostly for reading the file, so moving the preprocessing itself to the GPU would save at most about 0.5\,s.

\paragraph{The folder.} On the folder, ours pays start-up, model loading and warm-up once (3.3--3.7\,s on one scan in the integrations, 4.6--4.8\,s through
our command line for \nnunet), runs its
network compiled (a forward pass 26\,\% faster than eager) and reads and writes while the GPU works: through our command
line the 14 network passes follow each other with 0.02--0.10\,s of gaps in total, in the TotalSegmentator integration with
1.9--2.0\,s (0.14\,s per scan; the MOOSE integration runs as a subprocess and records no such events). TotalSegmentator 2.18 and the development branch load their
five networks again for every scan. MOOSE's folder mode pauses for 13\,s per scan (2\,s after each of its six models and 1\,s
more), 19\,\% of its 68\,s per scan; without these pauses ours would be about 12$\times$ faster on the folder instead of
15$\times$. Its single-scan path has no pauses.

\paragraph{What step 0.8 buys ours.} The same \TS-label model at \nnunet's step 0.5, through our command line, takes
1.7--1.8$\times$ as long as TotalSegmentator with ours at 0.8 on the GPU (10.5--20.0\,s against 5.8--11.3\,s) and 2.3--2.9$\times$
as long on the CPU. At the same step the command line takes 1.04--1.08$\times$ as long as the MOOSE integration on the
GPU and 1.09--1.10$\times$ on the CPU, so most of the difference is the step, not the integration. Since step 0.8 costs no measurable segmentation quality (Sec.~\ref{sec:supp_export}), our TotalSegmentator integration keeps it.
\begin{table}[h]
\caption{\textbf{The toolkits end to end, per scan} (Tab.~3 of the main paper; ranges there). GPU: median time of three runs,
largest peak memory. CPU: one run per scan; $^{\S}$from an earlier run with the same code and machine.}
\label{tab:release_scans}
\centering
\small
\resizebox{\linewidth}{!}{%
\begin{tabular}{@{}lcccc@{}}
\toprule
 & AutoPET (0371) & Leipzig (1046) & Careggi (1412) & UC Davis (1600) \\
\midrule
\multicolumn{5}{@{}l}{\textit{GPU: time (s) / RAM (GB) / VRAM (GB)}} \\
\TS 2.18 (5 networks, step 0.8) & 69.3 / 12.9 / 9.6 & 85.4 / 15.8 / 11.8 & 96.5 / 18.8 / 14.3 & 162 / 31.5 / 20.7 \\
\TS dev.\ branch (5, step 0.8) & 21.1 / 8.1 / 9.8 & 27.4 / 9.0 / 12.1 & 26.6 / 9.8 / 14.6 & 51.9 / 18.6 / 21.0 \\
\rowcolor{oursrow} \TS with ours (1, step 0.8) & 5.8 / 2.7 / 6.1 & 5.8 / 3.0 / 6.5 & 7.7 / 2.5 / 7.2 & 11.3 / 6.0 / 7.1 \\
\rowcolor{oursrow} \TS with ours, its 5 released (step 0.8) & 12.6 / 4.2 / 2.9 & 17.4 / 4.3 / 3.2 & 15.3 / 4.5 / 3.5 & 34.2 / 9.2 / 4.5 \\
MOOSE 3.2.2 (6, step 0.5) & 67.4 / 13.0 / 6.8 & 73.9 / 16.0 / 8.2 & 69.8 / 11.6 / 4.5 & 115 / 25.1 / 8.5 \\
\rowcolor{oursrow} MOOSE with ours (1, step 0.5) & 9.8 / 3.0 / 6.1 & 9.7 / 2.9 / 6.6 & 12.1 / 2.5 / 7.3 & 17.7 / 7.8 / 7.2 \\
\rowcolor{oursrow} MOOSE with ours, its 6 released (step 0.5) & 28.7 / 4.7 / 3.6 & 28.7 / 4.9 / 3.9 & 46.1 / 5.1 / 4.1 & 59.5 / 11.1 / 4.7 \\
\nnunet 2.8.1 (1, step 0.5, either model) & \multicolumn{4}{c}{\textit{out of RAM}} \\
\rowcolor{oursrow} \nnunet with ours, \TS model (step 0.5) & 10.5 / 3.0 / 6.1 & 10.5 / 2.9 / 6.5 & 12.7 / 2.6 / 7.3 & 20.0 / 6.4 / 7.2 \\
\rowcolor{oursrow} \nnunet with ours, \MOOSE model (step 0.5) & 10.3 / 3.0 / 6.1 & 10.5 / 2.7 / 6.6 & 12.6 / 2.4 / 7.3 & 18.7 / 5.8 / 7.2 \\
\midrule
\multicolumn{5}{@{}l}{\textit{CPU only (16 cores, 32 threads): time (s) / RAM (GB)}} \\
\TS 2.18 & 510$^{\S}$ / 12.0 & 869$^{\S}$ / 14.7 & 765$^{\S}$ / 17.7 & 1630$^{\S}$ / 30.0 \\
\TS dev.\ branch & 475$^{\S}$ / 10.2 & 821$^{\S}$ / 11.9 & 718 / 15.2 & 1531$^{\S}$ / 27.2 \\
\rowcolor{oursrow} \TS with ours & 120 / 8.9 & 120 / 9.3 & 206 / 9.9 & 325 / 10.0 \\
MOOSE 3.2.2 & 1773$^{\S}$ / 13.3 & 1779$^{\S}$ / 14.9 & 2028$^{\S}$ / 11.8 & 3546$^{\S}$ / 21.4 \\
\rowcolor{oursrow} MOOSE with ours & 317 / 9.0 & 314 / 9.3 & 431 / 10.0 & 741 / 10.0 \\
\nnunet 2.8.1, \TS model & \multicolumn{4}{c}{\textit{out of RAM}$^{\S}$} \\
\rowcolor{oursrow} \nnunet with ours, \TS model & 345 / 9.0 & 343 / 9.4 & 470 / 10.0 & 811 / 10.0 \\
\rowcolor{oursrow} \nnunet with ours, \MOOSE model & 347 / 9.0 & 346 / 9.4 & 470 / 10.0 & 812 / 10.1 \\
\bottomrule
\end{tabular}%
}
\end{table}

\paragraph{Where the gap comes from.}
Table~\ref{tab:attribution} splits the times of Table~\ref{tab:release_scans} into their parts; a toolkit and our integration of it
run at the same step size, 0.8 for TotalSegmentator and 0.5 for MOOSE, so both sides of each gap use the same step.
Of the 64--150\,s by which \TS 2.18 is slower than \TS with ours, 39--94\,s (61--67\,\%) are logits and label maps, 5.4--18\,s
(8--13\,\%) forward passes (five plain U-Nets with 315--1040 windows against our ResEnc U-Net with 16--36), and the remaining
19--39\,s (25--30\,\%) start-up, loading, reading, preprocessing and other, 17--35\,s of it reading and preprocessing.
The \TS development branch reads, preprocesses and handles its logits faster: of its 15--41\,s gap, 35--47\,\% are forward passes,
18--23\,\% logits and 34--44\,\% the rest, mostly the time between its networks, which also holds part of its handling of the
logits.
Of the 58--97\,s by which MOOSE is slower than MOOSE with ours, 24--39\,s (40--52\,\%) are logits and label maps, 21--40\,s
(33--42\,\%) forward passes (six plain U-Nets with 1224--2448 windows against our 54--117), and 8.3--17\,s (14--20\,\%) the rest.
The forward passes still combine the number of networks and their size; Tab.~1 of the main paper isolates the method at a common
step of 0.5, and its split (Table~\ref{tab:breakdown}) separates the two: with the toolkits' own plain U-Net, one network instead of
five or six saves 12--39\,s (\TS) and 21--42\,s (MOOSE) of forward passes, and our larger ResEnc U-Net instead of the plain one
costs ours 0.1--1.8\,s and 0.9--3.0\,s.
\begin{table}[h]
\caption{\textbf{Where the end-to-end time goes} (GPU, one scan): the times of Table~\ref{tab:release_scans} split into parts, in s,
ranges over the four scans, each part the median of three runs (the parts add up to the total within 1.1\,s). A toolkit and our
integration of it run at the same step size. Start-up and loading: interpreter, imports, CUDA context, model loading and exit; ours
also warms up the GPU here. Forward passes: GPU time of the networks' forward passes; networks and windows per scan in parentheses.
Logits to label map: everything in the network phase besides the forward passes (blending into the logit buffer; ours:
its fused conversion), each network's conversion to a label map, and everything after the last network (nearest-neighbor
upsampling, merging, undoing cropping and reorientation). Unlike in Tab.~1 of the main paper, this includes the transfer of the
logits to export processes (2.18, MOOSE 3.2.2) and writing; the toolkits write inside these steps, so writing is not separated (ours
writes in at most 0.6\,s). Other: time between the networks that the measurement does not attribute (ours runs one network).
$^\dagger$Read and preprocessed in a background thread while the model loads (reading alone takes 0.3--3.1\,s);
given is the main thread's time before its pass besides loading (on UC Davis, it waits for that thread). $^\ddagger$Between its networks, the development branch also moves each network's logits to the host
and takes their argmax.}
\label{tab:attribution}
\centering
\small
\setlength{\tabcolsep}{4pt}
\resizebox{\linewidth}{!}{%
\begin{tabular}{@{}llcccccc@{}}
\toprule
 & networks & start-up, & reading, & forward & logits to & & \\
pipeline & (windows) & loading & preprocessing & passes & label map, writing & other & total \\
\midrule
\multicolumn{8}{@{}l}{\textbf{\TS labels (117 classes), step 0.8}} \\
\quad release 2.18 & 5 (315--1040) & 5.0--5.2 & 17--37 & 7.0--21 & 40--96 & 0.6--2.2 & 69--162 \\
\quad dev.\ branch (Oct.\ 2026) & 5 (315--1040) & 5.1--5.3 & 1.0--4.4 & 6.9--21 & 4.2--11 & 3.6--9.9$^\ddagger$ & 21--52 \\
\rowcolor{oursrow} \quad with ours & 1 (16--36) & 3.3--3.5 & 0.1--2.3$^\dagger$ & 1.6--3.4 & 0.7--2.1 & -- & 5.8--11 \\
\midrule
\multicolumn{8}{@{}l}{\textbf{\enh labels (125 classes, MOOSE), step 0.5}} \\
\quad MOOSE 3.2.2 & 6 (1224--2448) & 6.1--7.7 & 3.6--8.8 & 26--51 & 24--41 & 1.0--6.8 & 67--115 \\
\rowcolor{oursrow} \quad with ours & 1 (54--117) & 3.6--3.7 & 0.3--1.2$^\dagger$ & 5.0--11 & 0.8--2.1 & -- & 9.7--18 \\
\bottomrule
\end{tabular}%
}
\end{table}

\section{TotalSegmentator development branch (October 2026)}
\label{sec:supp_tsv3}
Besides its releases, TotalSegmentator has a public development branch with provisional weights. We use
the code at commit \texttt{9885322} of 16 October 2026
(\url{https://github.com/wasserth/TotalSegmentator/commit/9885322}) and the weights tagged
\texttt{v3.0.0-weights} on 7 September 2026 (\url{https://github.com/wasserth/TotalSegmentator/releases/tag/v3.0.0-weights}). The
branch is not a release: its maintainer marks the weights as pre-release, and the branch may change or be withdrawn. The main
paper therefore compares with the latest release, 2.18, and we report the development branch here. It keeps the split into five plain U-Nets and the nearest-neighbor upsampling of their label maps, but it
already removes much of the overhead around the networks: it holds the images in memory and resamples the input on the GPU. Ours is
still 3.8--5.8$\times$ faster in prediction (Table~\ref{tab:toolkits_scans}), and end to end 3.5--4.7$\times$ faster on one scan,
8.4$\times$ on a folder and 3.5--6.9$\times$ on the CPU, with 3.0--4.0$\times$ less RAM (Table~\ref{tab:release_scans}; also
Tables~\ref{tab:step08} and~\ref{tab:breakdown}). The gain of our inference therefore goes beyond the overhead that the development branch removes.

\section{Training cost vs.\ the class count}
\begin{figure}[t]
\centering
\includegraphics[width=0.6\linewidth]{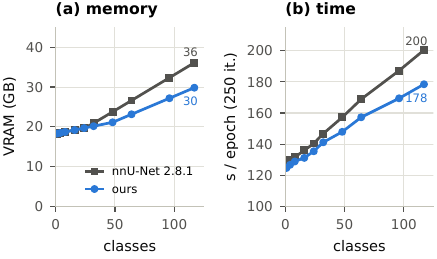}
\caption{\textbf{Training cost vs.\ the class count,} training step only: peak VRAM (a) and time per epoch of 250
iterations (b), short trainings on one L40S at 192$^3$, batch 2, \TS labels merged into $N$ classes. At 117 classes ours needs
30\,GB and 178\,s, \nnunet 2.8.1 36\,GB and 200\,s, so ours is 12\,\% faster.}
\label{fig:traincost}
\end{figure}

Figure~\ref{fig:traincost} backs Sec.~4 of the main paper: the training step of \nnunet 2.8.1 and ours at 192$^3$, batch 2.
\nnunet 2.8.1's validation step runs out of a 48\,GB GPU from 96 classes, so a whole 117-class training does not complete.
Ours counts the argmax of each validation patch into a $C{\times}C$ confusion matrix instead of building one-hot predictions
(Table~\ref{tab:cost}). With
only that step replaced, a whole \pgps training run of ours is 1.27--1.34$\times$ faster (L40S / A100), from the changes of Sec.~4
together with fused SGD, one host synchronization per epoch and CUDA graphs for small patches.

\paragraph{Training compute.} The GPU-hours of Sec.~4 of the main paper count training only, on one L40S. A dense 1000-epoch
run of our ResEnc U-Net at 192$^3$ takes 48.4 and 48.9 GPU-hours on \TS and \enh, measured per \pgps stage, and \pgps 5400
costs 1.64 times as much, 79 and 80 GPU-hours. The specialists, the toolkits' plain U-Nets (\TS: patch 128$^3$, 1000 epochs;
MOOSE: 224$\times$96$\times$96, 2000 epochs), take 42 and 102 GPU-hours together, from their measured time per iteration and the
iterations of their schedules.

\section{How the network and the schedule were chosen}
Our architecture follows the residual-encoder U-Net of nnU-Net Revisited~\cite{isensee2024revisited}. Its presets M, L and XL
target increasing GPU memory. For whole-body CT, the L preset, which fits a 24\,GB GPU, plans a patch of about 192$^3$ voxels at
batch size 2, and it is among the strongest configurations of that study. The schedule was chosen by the 5-fold CV of Sec.~S9. This section tests the network and its patch size with the same CV.
It starts from the toolkits' own setting and changes one factor at a time (Table~\ref{tab:recipe_cv}).

\paragraph{Setup.}
Folds, scoring and intervals are those of Sec.~S9. Each label space is split into five folds inside its training split, so no test
scan is used, and every arm and fold is one training run. Arms are compared by paired differences over the five folds with 95\,\%
$t$-intervals. The toolkits' setting is a single plain U-Net, the specialists' architecture with 31\,M parameters, trained for all
classes of the label space at a toolkit's patch size and schedule: 128$^3$ and 1000 dense epochs for \TS, 224$\times$96$\times$96
and 2000 dense epochs for MOOSE. From there, the arms change the schedule to \pgps 5400, the patch size to 192$^3$ and the network to
our ResEnc U-Net. On \enh, the plain U-Net's last pooling step changes with the patch shape, as \nnunet plans it. \pgps grows the
patch to each arm's own final size. All arms share our trainer, the data and the splits.

\begin{table}[h]
\caption{\textbf{From the toolkits' setting to our recipe.} 5-fold CV inside each training split (as Table~\ref{tab:cv_ladder}),
mean \Dsc over all classes, FP-penalized and on present structures; one training run per arm and fold. Every model covers all
classes of its label space. Toolkit's patch size: 128$^3$ (\TS), 224$\times$96$\times$96 (MOOSE on \enh); toolkit's schedule:
1000 (\TS) and 2000 (MOOSE) dense epochs. Shaded: our recipe, used for all our models.}
\label{tab:recipe_cv}
\centering
\footnotesize
\setlength{\tabcolsep}{5pt}
\begin{tabular}{@{}lllcccc@{}}
\toprule
 & & & \multicolumn{2}{c}{\TS (117 classes)} & \multicolumn{2}{c}{\enh (125)} \\
\cmidrule(lr){4-5}\cmidrule(l){6-7}
network & patch size & schedule & FP-pen. & present & FP-pen. & present \\
\midrule
plain U-Net & toolkit's & toolkit's & 85.14 & 87.83 & 84.76 & 88.80 \\
plain U-Net & toolkit's & \pgps 5400 & 86.34 & 88.58 & 86.53 & 89.27 \\
plain U-Net & 192$^3$ & \pgps 5400 & 87.36 & 89.45 & 87.59 & 90.03 \\
ResEnc U-Net & 192$^3$ & toolkit's & 86.59 & 89.61 & 87.86 & 90.15 \\
ResEnc U-Net & 128$^3$ & \pgps 5400 & 87.13 & 89.11 & 87.51 & 89.83 \\
ResEnc U-Net & 160$^3$ & \pgps 5400 & 87.78 & 89.75 & 87.87 & 90.07 \\
\rowcolor{oursrow} ResEnc U-Net & 192$^3$ & \pgps 5400 & 87.93 & 89.86 & 88.20 & 90.23 \\
\midrule
\multicolumn{7}{@{}l}{\textit{paired difference over the five folds [95\,\% $t$-interval]}} \\
\multicolumn{3}{@{}l}{plain U-Net, toolkit's patch size: \pgps 5400 $-$ toolkit's schedule} & +1.20 [1.03, 1.37] & +0.75 [0.61, 0.90] & +1.77 [1.24, 2.30] & +0.48 [$-$0.52, 1.48] \\
\multicolumn{3}{@{}l}{plain U-Net, \pgps 5400: 192$^3$ $-$ toolkit's patch size} & +1.01 [0.77, 1.26] & +0.87 [0.47, 1.26] & +1.05 [0.59, 1.52] & +0.75 [0.51, 1.00] \\
\multicolumn{3}{@{}l}{192$^3$, \pgps 5400: ResEnc $-$ plain U-Net} & +0.57 [0.51, 0.63] & +0.41 [0.29, 0.53] & +0.61 [0.42, 0.81] & +0.21 [0.05, 0.37] \\
\multicolumn{3}{@{}l}{ResEnc U-Net, \pgps 5400: 160$^3$ $-$ 128$^3$} & +0.65 [0.51, 0.78] & +0.64 [0.46, 0.82] & +0.36 [0.19, 0.54] & +0.25 [0.04, 0.46] \\
\multicolumn{3}{@{}l}{ResEnc U-Net, \pgps 5400: 192$^3$ $-$ 160$^3$} & +0.14 [0.03, 0.26] & +0.11 [$-$0.15, 0.37] & +0.33 [0.27, 0.39] & +0.16 [$-$0.03, 0.34] \\
\bottomrule
\end{tabular}
\end{table}

\paragraph{From the toolkits' setting to ours.}
On \TS, the plain U-Net at the toolkit's setting reaches 85.14 \Dsc points FP-penalized and 87.83 on present structures. On \enh it reaches 84.76 and 88.80. Trained with \pgps 5400 at the same patch size, it reaches 86.34 and 88.58 on \TS and 86.53 and 89.27
on \enh. With a patch size of 192$^3$, it reaches 87.36 and 89.45 on
\TS and 87.59 and 90.03 on \enh. Our ResEnc U-Net at the same patch size and schedule reaches 87.93 and
89.86 on \TS and 88.20 and 90.23 on \enh. FP-penalized, our recipe is therefore 2.79 points
above the toolkits' setting on \TS and 3.44 on \enh. On both label spaces, the patch size adds about one point FP-penalized
and the ResEnc U-Net about 0.6. FP-penalized, the schedule adds 1.2 on \TS and 1.8 on \enh. On present structures it adds 0.75 on \TS and 0.48 on \enh, where the gain varies strongly between folds. This comes from the toolkit's schedule, whose results on \enh vary much more across folds (standard deviation 0.95 against 0.29 with \pgps 5400; on \TS 0.42 against 0.38). The longer schedule thus also appears to make training more stable. With the toolkits' schedule instead of \pgps 5400, our network reaches 86.59 on \TS and 87.86
on \enh FP-penalized (Table~\ref{tab:cv_ladder}).

\paragraph{Patch size.}
With the ResEnc U-Net and \pgps 5400, the FP-penalized \Dsc rises from 87.13 at 128$^3$ to 87.78 at 160$^3$ and 87.93 at 192$^3$ on
\TS, and from 87.51 to 87.87 and 88.20 on \enh. Both steps gain FP-penalized on both label spaces, and on present
structures 160$^3$ is ahead of 128$^3$ as well. %
192$^3$ is slightly ahead of 160$^3$ on both label spaces, by 0.14 and 0.33 points FP-penalized and by 0.11 and 0.16 on present
structures. The intervals of the latter two contain 0. We therefore use 192$^3$.

\paragraph{Plain vs.\ ResEnc U-Net.}
At the same patch size and schedule, the ResEnc U-Net (102\,M parameters) is ahead of the plain U-Net by 0.57 \Dsc points
FP-penalized (95\,\% $t$-interval 0.51--0.63) and 0.41 on present structures (0.29--0.53) on \TS, and by 0.61 (0.42--0.81) and 0.21
(0.05--0.37) on \enh.

\section{Long schedules: the CV ladder}
\begin{figure}[t]
\centering
\includegraphics[width=0.67\linewidth]{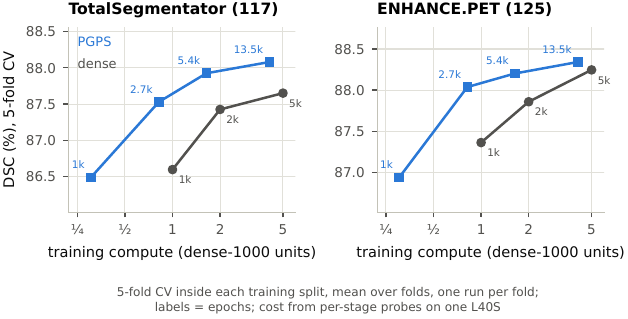}
\caption{\textbf{Many-class models need long schedules; progressive growing of patch size makes them affordable.} 5-fold CV inside each
training split, FP-penalized \Dsc, dense schedules (gray) vs.\ progressive growing of patch size (\pgps, blue) along their
epoch ladders, against training compute (one unit = a dense 1000-epoch run, from per-stage cost probes on one L40S).}
\label{fig:frontier}
\end{figure}
\paragraph{Setup.}
Each label space is split into five folds inside its training split, so no test scan is used: \TS 1,464 / 366 scans per fold within
the official 1,830, \enh 1,076--1,077 / 269--270 within our 1,346 (stratified by center and diagnosis like the test split). %
All arms train the recipe of the main paper (ResEnc U-Net, patch 192$^3$, batch 2, one GPU, no
mirroring) and differ only in the schedule: dense training for 1000, 2000 or 5000 epochs, or \pgps for 1000, 2700, 5400 or 13500
epochs. Every arm and fold is one training run with one seed, 70 runs in all. %
We score each fold's validation scans with the mean \Dsc over all classes, FP-penalized and on present structures, and compare arms
by paired differences over the five folds with 95\,\% $t$-intervals (4 degrees of freedom). Because every fold is a separate run,
these intervals contain run-to-run variation as well as the split's; they are the noise estimate of this section, and a difference
whose interval contains 0 is undistinguished. Table~\ref{tab:cv_ladder} lists the means and the key differences.

\begin{table}[h]
\caption{\textbf{The CV ladder.} 5-fold CV inside each training split, mean \Dsc over all classes, FP-penalized and on present
structures; one training run per arm and fold. Compute in units of a dense 1000-epoch run of the same configuration (cost probes on
one L40S). $^\dagger$: within the selection tolerance of the best FP-penalized arm (text);
shaded: the selected schedule, used for all our models.}
\label{tab:cv_ladder}
\centering
\footnotesize
\setlength{\tabcolsep}{5pt}
\begin{tabular}{@{}llcccc@{}}
\toprule
 & & \multicolumn{2}{c}{\TS (117 classes)} & \multicolumn{2}{c}{\enh (125)} \\
\cmidrule(lr){3-4}\cmidrule(l){5-6}
arm & compute & FP-pen. & present & FP-pen. & present \\
\midrule
dense 1000 & 1.00 & 86.59 & 89.61 & 87.36 & 90.08 \\
dense 2000 & 2.00 & 87.43 & 89.87 & 87.86 & 90.15 \\
dense 5000 & 5.00 & 87.65 & 89.89 & 88.25$^\dagger$ & 90.29 \\
\pgps 1000 & 0.30 & 86.48 & 89.20 & 86.94 & 89.79 \\
\pgps 2700 & 0.82 & 87.53 & 89.70 & 88.04 & 90.23 \\
\rowcolor{oursrow} \pgps 5400 & 1.64 & 87.93$^\dagger$ & 89.86 & 88.20$^\dagger$ & 90.23 \\
\pgps 13500 & 4.09--4.10 & 88.08$^\dagger$ & 89.88 & 88.34$^\dagger$ & 90.29 \\
\midrule
\multicolumn{6}{@{}l}{\textit{paired difference over the five folds [95\,\% $t$-interval]}} \\
\multicolumn{2}{@{}l}{dense 2000 $-$ dense 1000} & +0.83 [$-$0.20, 1.86] & +0.26 [$-$0.11, 0.64] & +0.50 [$-$0.03, 1.02] & +0.07 [$-$0.03, 0.17] \\
\multicolumn{2}{@{}l}{\pgps 1000 $-$ dense 1000} & $-$0.11 [$-$0.99, 0.77] & $-$0.41 [$-$0.72, $-$0.09] & $-$0.43 [$-$0.85, $-$0.00] & $-$0.29 [$-$0.43, $-$0.16] \\
\multicolumn{2}{@{}l}{\pgps 5400 $-$ dense 2000} & +0.50 [0.27, 0.73] & $-$0.01 [$-$0.29, 0.26] & +0.34 [0.12, 0.57] & +0.08 [0.05, 0.11] \\
\multicolumn{2}{@{}l}{dense 5000 $-$ \pgps 5400} & $-$0.27 [$-$0.48, $-$0.07] & +0.03 [$-$0.10, 0.16] & +0.04 [$-$0.09, 0.17] & +0.05 [$-$0.02, 0.13] \\
\multicolumn{2}{@{}l}{\pgps 13500 $-$ \pgps 5400} & +0.16 [0.02, 0.29] & +0.02 [$-$0.13, 0.17] & +0.14 [0.06, 0.22] & +0.06 [$-$0.00, 0.12] \\
\bottomrule
\end{tabular}
\end{table}

\paragraph{Results.}
Dense training keeps improving beyond 1000 epochs: 2000 epochs add 0.83 / 0.50 \Dsc points FP-penalized (\TS{} / \enh),
positive in 4 and 5 of the 5 folds, though both intervals contain 0; on present structures they add 0.26 / 0.07. %
Most of the FP-penalized gain is fewer predictions of structures a scan does not contain; without predictions under 50 voxels
(0.17\,ml) it is 0.35 / 0.26. %
\pgps is weaker per epoch (at 1000 epochs it trails dense training by 0.29--0.41 points on present structures) but is ahead
FP-penalized at about equal compute in all six pairs (\pgps 2700, 5400 and 13500 against dense 1000, 2000 and 5000). %
At a third of the compute, \pgps 5400 lies 0.27 [0.07, 0.48] points above dense 5000 on \TS, in all five folds (dense 5000 makes 534
instead of 464 predictions of absent structures per fold), and 0.04 below it on \enh (undistinguished); on
present structures the two are within 0.1 on both. %
\pgps 13500 adds 0.16 / 0.14 points for 2.5$\times$ the compute of \pgps 5400. %
On \TS this gain is fewer small false detections: without predictions under 50 voxels it is 0.02, and on present
structures it is +0.02. On \enh it is small but consistent (all five folds, 0.13 after the same filter), with the
largest per-class gains on hand and foot bones. %

\paragraph{Selection.}
We take the cheapest arm whose 5-fold mean FP-penalized \Dsc lies within max(2 paired s.e., 0.2 points) of the best arm's. The best
arm is \pgps 13500 on both label spaces. \pgps 5400 is 0.16 / 0.14 points below it, inside the tolerance of 0.20,
and every cheaper arm is outside (closest: \pgps 2700, 0.55 / 0.30 below), so the rule selects \pgps 5400 on
both. Dense 5000 is outside on \TS (0.43 below) and inside on \enh, at three times the compute of \pgps 5400. %

\paragraph{Compute.}
One unit on the compute axis of Fig.~\ref{fig:frontier} is a dense 1000-epoch run of the same configuration on the same GPU type:
48.4 / 48.9 L40S GPU-hours for \TS{} / \enh, training only; dense 2000 and 5000 cost 2 and 5 units. The runs
themselves trained on mixed GPU types, so their own clocks are not used. %
For \pgps we run the real trainer with its data loader for 8 epochs in every stage of the schedule (16 stages) on one
L40S, drop the first two epochs of each stage (compilation, CUDA-graph capture) and take the median of the other six. A run of $N$
epochs then costs $\sum_s n_s(N)\,t_s / (1000\,t_{\mathrm{full}})$ units, with $n_s(N)$ the epochs the schedule gives stage $s$,
$t_s$ its probed epoch time and $t_{\mathrm{full}}$ that of the full patch: 0.30, 0.82, 1.64 and 4.09--4.10 units
for \pgps 1000, 2700, 5400 and 13500. %
Against four production \pgps runs on \enh on the same node type, the probed epoch times agree within $-$5 to +1\,\% per stage;
production is 2--5\,\% faster in the three smallest stages and within 1\,\% from the fourth on. %

\section{Exactness and cost of streaming inference}
Proposition~1 of the main paper makes streaming exact in exact arithmetic. In floating point, streaming and the canonical pipeline
add the same terms in a different order and round in different places, so we compared their label maps voxel by voxel.

\paragraph{Setup.} We used 19 test scans: 10 from \TS's official test split, which is stored at the working spacing of 1.5\,mm
(no resampling), and 9 from the \enh test split at the scanner's resolution (10 drawn at random among scans with at most $10^8$
voxels, so that the canonical pipeline fits in RAM; on one it still did not). The networks are our ResEnc U-Nets for both label
sets with the head sliced to 32 classes, few enough for the canonical pipeline to run; with 33 output channels, streaming uses the
feature ring. The
canonical pipeline is \nnunet's default predictor with the same linear resampling. All variants of a scan ran in one process in a
fixed order, eager, with deterministic cuDNN, step 0.5 and no test-time augmentation. We report differing voxels as a share of the
voxels that either label map assigns to the foreground.

\paragraph{In fp32.} With the network in fp32 as well (no mixed precision, TF32 off) and fp32 buffers in both pipelines, the label
maps agree up to 0--3 voxels per scan and are identical on 10 of the 19 scans. At every differing voxel the two highest reference
logits differ by at most $1.3\times10^{-5}$: these are near-ties, which the other order of summation breaks the other way.

\paragraph{As deployed.} Networks run in mixed precision and emit fp16. The canonical pipeline blends their fp16 logits in an fp16
buffer; ours blends fp16 features and applies the head in fp32. The label maps of the two then differ in 0.21\,\% (\TS) and
0.096\,\% (\enh) of the foreground (medians over scans; at most 0.30\,\%). Against the all-fp32 result, streaming is the closer of
the two on each of the 19 scans: its label map differs in 0.12\,\% and 0.050\,\% of the foreground, the canonical one in 0.20\,\%
and 0.096\,\%, 1.7 and 1.9 times as much. The gain comes from where the head runs, not from the buffers: with fp32 buffers in
both pipelines the canonical label map is still 0.12\,\% and 0.069\,\% away, ours 0.065\,\% and 0.038\,\%, the rest being the
mixed precision inside the network, common to both. For scale, letting cuDNN choose other convolution
algorithms changes the canonical label map by 0.066\,\% and 0.038\,\%. The mean \Dsc over classes of every variant lies within
0.022 points of the all-fp32 result on every scan.

\paragraph{Feature ring vs.\ logit ring.} Our predictor can also keep the logits in the ring, one channel per class, and does
so where they need no more channels than the 32 features. We compared both rings on three \TS and three \enh test scans, each with
two of our ResEnc U-Nets (160$^3$ and 192$^3$ windows, 117 and 125 classes). With everything in fp32 the label maps are identical
on 7 of the 8 pairs on which the fp32 logit ring fit in VRAM and differ in one voxel on the eighth; with test-time mirroring along
two axes they differ in 0--2 voxels on five scans. In fp16 they differ in 0.017--0.040\,\% of the foreground, and on each of the
8 pairs the feature ring is the closer to the all-fp32 result, by 1.8--3.4$\times$, again because it applies the head in fp32.
Table~\ref{tab:ablation} gives the cost of either ring.

\paragraph{Compiled vs.\ eager.} Our predictor compiles the network when it predicts more than four scans in one call; the
comparisons of Tab.~1 and Fig.~3 of the main paper run eager. Compilation changes the label map slightly: on the 12 pairs above by
87--499 voxels with the feature ring (0.005--0.011\,\% of the foreground) and by 148--820 with the logit ring (0.009--0.020\,\%),
and on the whole-body \enh scan 0545 (97.8 million voxels) at 125 classes by 377 voxels. Two compilations of the same network need
not agree either: with an MRI model (MRISegmentator, 62 classes), separate processes gave label maps 38--65 voxels apart per scan
(of about 9 million), whereas eager runs were identical across processes. These differences come from the compiled kernels, not
from streaming: the logit ring changes more under compilation, on every pair, and the two rings differ by the same amount compiled
as eager (0.018--0.044\,\% vs.\ 0.017--0.040\,\% of the foreground).

\paragraph{Ensembles.} For an ensemble of $K$ models the ring holds the members' features side by side, $KF$ channels, and every
finished slice gets each member's head. If the members' logits are averaged, as \nnunet does over the folds of a configuration, the
heads stack into one $C\times KF$ matrix $\frac1K[W_1\,\cdots\,W_K]$ with bias $\frac1K\sum_k b_k$, and one product per slice gives
the mean logits; where $KF\geq C$, the ring holds the $C$ averaged logits instead, as they are then the smaller buffer (at 117 or 125
classes from four members on). Averaging probabilities, as \nnunet does across configurations, places the softmax between
resampling and averaging, so the members' heads cannot be merged: each member's logits are resampled before its softmax, which
Proposition~1 permits member by member. Our predictor implements logit averaging only. We measured it with two- and three-member
ensembles of \TS ResEnc U-Nets (160$^3$ windows, 117 classes; checkpoints from two training runs) on three \TS test scans. With
everything in fp32 the feature ring and the logit ring give identical label maps in all six cases; in fp16 the feature ring is
0.005--0.010\,\% of the foreground away from the all-fp32 result and the logit ring 0.024--0.044\,\%. On the largest scan, the peak
memory PyTorch allocates drops from 6.2\,GB with the logit ring to 4.5\,GB with two members and 5.5\,GB with three.

\paragraph{What each step buys.} Table~\ref{tab:ablation} separates the two steps on the head-to-toe scan of Fig.~3 of the main
paper. With the same network and still one ring channel per class, ours needs 7.8\,s, 2.1\,GB of RAM and 2.9\,GB of VRAM at 10
classes where \nnunet's default needs 68.2\,s (60.3\,s of it converting the logits on the CPU), 23.3\,GB and 9.7\,GB. Our whole
step takes less time than the canonical sliding-window prediction alone (8.2\,s), so streaming costs nothing at few classes.
Moving the head behind blending then takes the class dimension out of the ring: at 125 classes VRAM drops from 10.9
to 6.2\,GB and time by 12\,\%, and from 40 classes on VRAM stays at 6.1--6.2\,GB while the logit ring grows with every class. On the
four scans of
Tab.~1 of the main paper (three interleaved runs each, both label sets, same protocol as Table~\ref{tab:ablation}b), the feature
ring needs 5.4--6.5 instead of 8.8--12.8\,GB of VRAM and takes 11--14\,\% less time, at the same RAM.
\begin{table}[h]
\caption{\textbf{What each step of our inference buys}, on the head-to-toe scan of Fig.~3 of the main paper (\enh 0484) at 10, 25
and 125 classes (classification head sliced, as in the class sweep there): time in s, peak RAM and VRAM in GB; step 0.5, no
test-time augmentation, one run per point.
(a)~Prediction as in Fig.~3 and Tab.~1 of the main paper (Sec.~\ref{sec:supp_measure}): \nnunet's default as
released, which resamples the logits on the CPU, and ours, both on the same plain U-Net. (b)~Our predictor on our ResEnc U-Net with
the logit ring forced and as shipped, timed as in (a) but with the network compiled (compilation outside the timed window).
The two blocks differ in network and compilation; compare rows within a block only. $^\dagger$As shipped, the ring holds the 32 features, or the logits where these need no more
channels (10 and 25 classes); there the two rows of (b) run the same configuration. $^\ddagger$Out of RAM (55\,GB) already at 40
classes; larger counts were not run.}
\label{tab:ablation}
\centering
\footnotesize
\setlength{\tabcolsep}{3pt}
\begin{tabular}{l*{9}{N{0.95cm}}}
\toprule
 & \multicolumn{3}{c}{10 classes} & \multicolumn{3}{c}{25 classes} & \multicolumn{3}{c}{125 classes} \\
\cmidrule(lr){2-4}\cmidrule(lr){5-7}\cmidrule(lr){8-10}
 & time & RAM & VRAM & time & RAM & VRAM & time & RAM & VRAM \\
\midrule
\multicolumn{10}{l}{\textbf{(a) Canonical vs.\ streaming}: prediction, eager; the same plain U-Net in both} \\
\nnunet 2.8.1 default (canonical) & 68.2 & 23.3 & 9.7 & 185.6 & 49.2 & 20.1 & \multicolumn{3}{c}{out of RAM$^\ddagger$} \\
\rowcolor{oursrow} \textbf{ours}, as shipped$^\dagger$ & 7.8 & 2.1 & 2.9 & 8.1 & 2.1 & 3.6 & 8.5 & 2.0 & 4.1 \\
\midrule
\multicolumn{10}{l}{\textbf{(b) Head behind blending, per finished slice}: prediction as in (a), compiled; our ResEnc U-Net} \\
ours, logit ring (one channel per class) & 8.2 & 2.4 & 5.2 & 8.4 & 2.4 & 5.8 & 10.1 & 2.4 & 10.9 \\
\rowcolor{oursrow} \textbf{ours}, as shipped$^\dagger$ & 8.1 & 2.4 & 5.2 & 8.4 & 2.4 & 5.8 & 8.8 & 2.5 & 6.2 \\
\bottomrule
\end{tabular}
\end{table}

\section{Nearest-neighbor vs.\ exact export}
\label{sec:supp_export}
Sec.~5.2 of the main paper exports the same network output in two ways. Here we give the setup, the selection of the AMOS scans, the
results per class and the normalized surface Dice.

\paragraph{Exports.} For each scan, our predictor runs the network on the preprocessed image (1.5\,mm working grid, step 0.5 unless stated,
no test-time augmentation) and exports its output three times. \emph{Exact} (ours): the logits are resampled linearly to the scanner's grid
(\texttt{align\_corners=False}) and the argmax is taken there. \emph{Nearest neighbor}: the argmax is taken on the working grid and
the label map is upsampled with \texttt{nearest-exact} interpolation, which aligns the two grids at their voxel centers, as the
resampling of the input does. \emph{Zoom}: the same label map upsampled with scipy's \texttt{zoom} of order 0, TotalSegmentator
2.18's label upsampling. It maps the centers of the outermost voxels onto each other instead, which adds a sub-voxel shift, so
nearest-exact is the clean contrast. All three label maps then undo cropping and reorientation in the same way and are scored against the reference with
the rules of the main paper. \Dsc is the mean over scans of each scan's mean over classes, and intervals are 95\,\% paired
case-bootstrap intervals. Each comparison uses one trained model on both sides (one training run per label space), so training
variance does not enter the differences.

\paragraph{Models and data.} The models are the single models of Tab.~2 of the main paper, the \TS-label model (117 classes) and the
\enh-label model (125 classes). Their classes are mapped to each dataset's classes by name; where our label space splits a structure,
the parts are merged (kidney and kidney cyst). Predicted classes without a counterpart count as background. BTCV: the 30 labeled scans
with manual labels of 13 abdominal organs, all 13 mapped for the \TS-label model and 12 for the \enh-label model, whose label space has
no esophagus; pixel spacing 0.59--0.98\,mm, slices 2.5--5\,mm. AMOS: the CTs of the post-challenge release with reference labels
(300 scans; its other labeled scans are MR), 14 of the 15 classes for the \TS-label model (the joint prostate/uterus class has no
counterpart) and 12 for the \enh-label model (no esophagus, no duodenum).

\paragraph{Selection of the AMOS scans.} Some AMOS reference labels were drawn on a coarser in-plane grid than the image: their
boundaries fall on every second pixel. Against such a reference, an export that places boundaries on the scanner's grid is penalized
for boundaries the reference cannot draw. We therefore keep the CTs whose labels follow the image's grid. Per axis, a block index
compares the number of label boundaries at even and at odd positions, $|n_\text{even}-n_\text{odd}|/(n_\text{even}+n_\text{odd})$, in
windows of 16 positions over the middle half of the axis, averaged over windows: it is near 0 for labels drawn on the image's grid and
near 1 when every boundary sits on one parity. A scan is kept if the index is below 0.2 on both in-plane axes. The criterion reads the
reference only, not the predictions. It keeps 255 of the 300 CTs (pixel spacing 0.45--1.07\,mm, slices 1.25--5\,mm) and excludes
amos\_0004, 0009, 0010, 0011, 0013, 0014, 0016, 0017, 0019, 0023, 0024, 0025, 0027, 0029, 0033, 0034, 0038, 0040, 0041, 0042, 0051,
0059, 0061, 0063, 0064, 0066, 0069, 0073, 0075, 0078, 0083, 0084, 0085, 0087, 0088, 0092, 0094, 0098, 0172, 0367, 0372, 0385, 0387,
0405 and 0410. In plane, one working voxel spans 1.4--3.3 pixels of the scanner's grid along each axis on the kept scans, 1.5--2.5 on
BTCV.

\paragraph{Results.} Nearest neighbor lowers the \Dsc of the \TS-label model on BTCV by 0.88 points (0.70--1.06; 30 of 30 scans lower)
and that of the \enh-label model by 0.47 (0.37--0.57; 28 of 30). On AMOS, the two models lose 0.66 (0.61--0.71; 244 of 255 scans
lower) and 0.52 (0.47--0.56; 227 of 255). Per class (Table~\ref{tab:export_perclass}), the 95\,\% interval of the loss lies below 0
for every class of both models on both datasets except the gallbladder of the \enh-label model on BTCV.
TotalSegmentator's zoom costs more: 1.29 and 0.65 points on BTCV (30 and 29 of 30
scans lower) and 1.06 (0.97--1.15) and 0.63 (0.57--0.69) on AMOS. On present structures only, every nearest-neighbor difference is the
same to within 0.01 points.

\paragraph{Relabeled foreground.} Independently of any reference, nearest neighbor gives 3.5\,\% of the foreground voxels a different
label than the exact export for the \TS-label model on BTCV (2.3--4.7\,\% per scan) and 2.8\,\% (2.2--3.5\,\%) for the \enh-label
model, and 3.5\,\% (2.2--4.4\,\%) and 3.0\,\% (2.2--4.1\,\%) on AMOS; the zoom relabels 4.2 and 3.0\,\% on BTCV and 4.7 and 3.5\,\% on
AMOS. Foreground is every voxel that either export assigns to one of the scored classes.

\paragraph{Normalized surface Dice.} The NSD of a class is the fraction of the two boundaries, the reference's and the
prediction's, that lies within a tolerance of the other (Sec.~5.2 of the main paper). We compute it with DeepMind's reference
implementation on the scanner's grid with its spacing and average it like \Dsc, with the same rule: a class absent from the
reference but predicted scores 0, a class absent from both is skipped. Table~\ref{tab:export_nsd} gives it at 1\,mm, the tolerance we
fixed before scoring, and at 0.7\,mm. At 1\,mm, nearest neighbor costs the \TS-label model 1.08 points on BTCV (0.59--1.52) and 0.88
on AMOS (0.75--0.99), and the \enh-label model 0.57 on AMOS (0.44--0.68); on BTCV the \enh-label model's NSD does not change (+0.04,
$-$0.22 to 0.31). At 0.7\,mm nearest neighbor costs 1.1--3.0 points, but this tolerance lies below the pixel spacing of many scans.
Surface distances on a voxel grid take discrete values (0, one pixel, a pixel diagonal, \dots). Where the in-plane pixel is 0.7\,mm or
larger, a 0.7\,mm tolerance accepts only boundary points on exactly the right pixel, and the loss at 0.7\,mm comes mostly from these
scans (Table~\ref{tab:export_nsd}, right). At 1\,mm, the loss is larger on the scans with finer pixels. The main paper quotes the
1\,mm values. On present structures only, every NSD difference is the same to within 0.02 points.

\paragraph{Step size.} With TotalSegmentator's step 0.8 instead of 0.5, the exact export changes the \Dsc on BTCV by +0.01 points
(\TS-label model, $-$0.02 to 0.04) and +0.03 (\enh-label model, 0.00 to 0.08). No class changes by more than 0.15 points, and 2 of the
25 per-class intervals exclude 0, about what chance alone gives at this many comparisons. Nearest neighbor costs as much at step 0.8
(0.88 and 0.44 points against the exact export at 0.5), and TotalSegmentator's recipe as a whole, step 0.8 with its zoom, costs our
models 1.30 (1.11--1.50) and 0.62 (0.49--0.75) points, about what the zoom alone costs.

\begin{table}[h]
\caption{\textbf{Nearest-neighbor vs.\ exact export per class,} \Dsc in points, FP-penalized, mean over scans; BTCV 30 scans,
AMOS 255 CTs. exact: our export at step 0.5. NN: nearest neighbor $-$ exact, the same network output. step 0.8: exact at step 0.8
$-$ exact at step 0.5. --: class not in the model's label space or not in the dataset. A scan without the structure scores 0 if the
model predicts it (BTCV gallbladder: two scans; AMOS: 8 and 13 (scan, class) pairs). Bottom rows: macro \Dsc as in Sec.~5.2 of the
main paper (mean over scans of each scan's mean over classes); TotalSegmentator 2.18's zoom $-$ exact; foreground voxels whose label
differs between nearest neighbor and exact.}
\label{tab:export_perclass}
\centering
\footnotesize
\setlength{\tabcolsep}{4pt}
\begin{tabular}{@{}lrrrrrrrrrr@{}}
\toprule
 & \multicolumn{6}{c}{BTCV} & \multicolumn{4}{c}{AMOS} \\
\cmidrule(lr){2-7}\cmidrule(l){8-11}
 & \multicolumn{3}{c}{\TS-label model} & \multicolumn{3}{c}{\enh-label model} & \multicolumn{2}{c}{\TS-label} & \multicolumn{2}{c}{\enh-label} \\
\cmidrule(lr){2-4}\cmidrule(lr){5-7}\cmidrule(lr){8-9}\cmidrule(l){10-11}
class & exact & NN & step 0.8 & exact & NN & step 0.8 & exact & NN & exact & NN \\
\midrule
spleen & 95.9 & $-$0.75 & 0.00 & 93.5 & $-$0.33 & +0.03 & 95.3 & $-$0.37 & 93.6 & $-$0.27 \\
right kidney & 92.0 & $-$0.81 & 0.00 & 91.6 & $-$0.40 & +0.01 & 95.1 & $-$0.64 & 93.2 & $-$0.51 \\
left kidney & 93.7 & $-$0.79 & 0.00 & 93.1 & $-$0.36 & $-$0.02 & 94.5 & $-$0.64 & 93.0 & $-$0.50 \\
gallbladder & 71.4 & $-$0.45 & +0.15 & 69.8 & +0.01 & +0.09 & 80.6 & $-$0.56 & 77.7 & $-$0.37 \\
esophagus & 78.1 & $-$0.79 & +0.05 & -- & -- & -- & 81.7 & $-$1.34 & -- & -- \\
liver & 97.1 & $-$0.43 & 0.00 & 96.8 & $-$0.18 & +0.03 & 96.4 & $-$0.22 & 95.8 & $-$0.17 \\
stomach & 93.0 & $-$0.40 & $-$0.01 & 88.4 & $-$0.12 & +0.04 & 89.2 & $-$0.32 & 87.6 & $-$0.20 \\
aorta & 93.0 & $-$1.09 & +0.02 & 93.0 & $-$0.90 & 0.00 & 93.1 & $-$1.25 & 92.1 & $-$0.91 \\
inferior vena cava & 88.0 & $-$0.71 & +0.02 & 85.9 & $-$0.71 & +0.01 & 83.3 & $-$0.43 & 80.3 & $-$0.38 \\
portal and splenic vein & 77.0 & $-$1.62 & $-$0.11 & 77.5 & $-$0.69 & +0.01 & -- & -- & -- & -- \\
pancreas & 82.3 & $-$0.61 & $-$0.05 & 82.3 & $-$0.29 & +0.04 & 81.1 & $-$0.51 & 80.7 & $-$0.38 \\
right adrenal gland & 72.9 & $-$1.70 & +0.06 & 72.3 & $-$0.87 & +0.10 & 71.4 & $-$1.59 & 70.3 & $-$1.35 \\
left adrenal gland & 69.8 & $-$1.34 & 0.00 & 69.9 & $-$0.75 & +0.06 & 69.8 & $-$1.00 & 70.3 & $-$0.97 \\
duodenum & -- & -- & -- & -- & -- & -- & 68.4 & $-$0.28 & -- & -- \\
urinary bladder & -- & -- & -- & -- & -- & -- & 80.0 & $-$0.11 & 81.2 & $-$0.22 \\
\midrule
macro \Dsc & 84.9 & $-$0.88 & +0.01 & 84.6 & $-$0.47 & +0.03 & 84.3 & $-$0.66 & 84.7 & $-$0.52 \\
\TS's zoom & & $-$1.29 & & & $-$0.65 & & & $-$1.06 & & $-$0.63 \\
relabeled (\%) & & 3.5 & & & 2.8 & & & 3.5 & & 3.0 \\
\bottomrule
\end{tabular}
\end{table}

\begin{table}[h]
\caption{\textbf{Normalized surface Dice (NSD) of the exact export and its change,} in points, FP-penalized, mean over scans;
BTCV 30 scans, AMOS 255 CTs. exact: our export. NN: nearest neighbor $-$ exact; zoom: TotalSegmentator 2.18's zoom $-$ exact; $^*$: the
95\,\% paired case-bootstrap interval excludes 0. NN, pixel: NN $-$ exact on the scans with an in-plane pixel spacing below 0.7\,mm
and of at least 0.7\,mm (BTCV 5 and 25 scans, AMOS 155 and 100).}
\label{tab:export_nsd}
\centering
\footnotesize
\setlength{\tabcolsep}{4pt}
\begin{tabular}{@{}lrrrrrrrrrr@{}}
\toprule
 & \multicolumn{5}{c}{tolerance 1\,mm} & \multicolumn{5}{c}{tolerance 0.7\,mm} \\
\cmidrule(lr){2-6}\cmidrule(l){7-11}
 & & & & \multicolumn{2}{c}{NN, pixel} & & & & \multicolumn{2}{c}{NN, pixel} \\
\cmidrule(lr){5-6}\cmidrule(l){10-11}
 & exact & NN & zoom & $<$0.7 & $\geq$0.7 & exact & NN & zoom & $<$0.7 & $\geq$0.7 \\
\midrule
BTCV, \TS-label model & 79.1 & $-$1.08$^*$ & $-$1.88$^*$ & $-$2.23 & $-$0.85 & 63.8 & $-$3.00$^*$ & $-$3.84$^*$ & $-$1.20 & $-$3.36 \\
BTCV, \enh-label model & 78.0 & +0.04 & $-$0.17 & $-$0.86 & +0.22 & 63.1 & $-$1.71$^*$ & $-$1.94$^*$ & +0.23 & $-$2.09 \\
AMOS, \TS-label model & 75.4 & $-$0.88$^*$ & $-$1.79$^*$ & $-$1.24 & $-$0.31 & 67.0 & $-$1.49$^*$ & $-$2.42$^*$ & $-$0.49 & $-$3.05 \\
AMOS, \enh-label model & 76.3 & $-$0.57$^*$ & $-$0.54$^*$ & $-$0.91 & $-$0.03 & 68.1 & $-$1.14$^*$ & $-$1.05$^*$ & $-$0.15 & $-$2.66 \\
\bottomrule
\end{tabular}
\end{table}

\section{Evaluation at 100+ classes}
All numbers in this section come from one training run per arm, the single model and one set of composed specialists per label
space; the intervals cover the test scans, not training variance.

\paragraph{Two rules from one count.} For every test scan and class we count the reference voxels, the predicted voxels and their
overlap, and compute \Dsc from these counts. A (scan, class) pair that neither the reference nor the prediction contains is not
scored. The FP-penalized rule (\nnunet's, our primary) scores every other pair, so a prediction of a structure that the scan does
not contain scores 0. The present-structure rule (FP-blind; the evaluators of TotalSegmentator and MOOSE) scores only the pairs whose
reference contains the class and ignores predictions of absent structures. Under both rules, a present structure that is not
predicted scores 0. The macro \Dsc averages each class over its scored scans and then all classes with equal weight.

\paragraph{Why we report both.} At 100+ classes a scan contains only part of the label space. On the test splits a scan contains 69
of the 117 \TS classes on average and 104 of the 125 \enh classes, so 41 and 17\,\% of the (scan,
class) pairs are absent structures. The present-structure rule measures how well present structures are delineated and is blind to
false detections. The FP-penalized rule scores each false detection like a structure missed entirely, as 0 for that pair, so it
also measures whether a model knows which structures a scan contains. Missed structures are rare for both arms (50 / 50 and 27 / 17
of 7,521 and 34,901 present pairs, single model / specialists), false detections are not (128 / 146 and 387 / 467
pairs), and so the rules can disagree: on \enh the single model is 0.44 points ahead FP-penalized and level on present
structures ($-$0.06; Tab.~2 of the main paper).

\paragraph{Slivers.} Most false detections are small. On \TS (1.5\,mm voxels) the median false detection has 34 voxels for the
single model and 32 for the specialists (about 0.1\,ml), and 69 and 65\,\% of them have fewer than 100 voxels (0.34\,ml); in
cross-validation they are mostly fragments of structures that the field of view cuts. On \enh the medians are 1,384 and 912
voxels. Under the FP-penalized rule a sliver costs as much as a missed organ.

\paragraph{Classes a model never segments.} The macro weighs every class equally, however few scans contain it. A class that a
model never segments scores 0 on every scan that contains it, under both rules, and is not scored on the other scans; even if it
occurs in a single test scan, it moves the macro by up to $100/C$ points (0.85 at $C = 117$). Rare classes therefore dominate small
macro differences, which is why we compare arms through paired case-bootstrap intervals rather than by their means.

\paragraph{Per class.} Table~\ref{tab:perclass} summarizes the per-class differences by specialist; about 5\,\% of the classes, six
of 117 or 125, would fall outside their 95\,\% intervals by chance alone. On \TS 15 classes are ahead and 9 behind FP-penalized (16
and 6 on present structures), on \enh 26 and 36 (24 and 51). Most differences are small: on present
structures 94 of the 117 \TS classes and 107 of the 125 \enh classes differ by less than one point. The
largest gains are on \TS's two kidney-cyst classes (+20.3 and +8.6 points FP-penalized), small structures present in 22 and 27 of
the 109 test scans. Without them the single model is +0.28 [$-$0.71, +0.64] points ahead FP-penalized and +0.25 [+0.04, +0.50] on
present structures.

\paragraph{Where the single model differs.} \emph{Merging.} The composed specialists claim the same voxel for two classes in
0.030\,\% (\TS) and 0.026\,\% (\enh) of their foreground, where the specialist merged last keeps it. Merge
conflicts are rare.

\emph{Boundaries the split cuts.} If the split hurt by hiding a class's neighbors from its specialist, the single model should gain
most on classes whose boundary touches the classes of other specialists. From the reference labels of the test splits (\enh: every
fifth scan, 68 of 337) we measure, for each class, the share of its boundary area that touches a class of another specialist. The
median share is 0.02 (\TS) and 0.01 (\enh); 3--4\,\% of a boundary touches classes of the same specialist, and
93--94\,\% touches unlabeled tissue. Across classes, the present-structure gain grows weakly with the share on \TS (Spearman +0.20
[+0.01, +0.33]) and moderately on \enh (+0.41 [+0.30, +0.46]); for the FP-penalized gain
the correlations are $-$0.22 [$-$0.29, +0.03] and +0.06 [$-$0.07, +0.20]. Within the classes of one
specialist the rank correlation is about 0 on both ($-$0.06 and +0.06 on present structures). The gain differs between
specialists, and the correlation across classes reflects which specialist a class belongs to, not how much of its boundary the split
cuts.

\emph{Specialists.} The clearest group is the ribs: on present structures the single model is ahead by 0.93 (\TS) and 0.90 (\enh)
points on average over the rib specialist's classes, with 10 of 26 and 22 of 27 of them ahead beyond their intervals and none behind.
On \TS the FP-penalized gain sits in the classes of the organ and muscle specialists (+1.90 and +1.22; organs without the kidney
cysts +0.77) and comes mostly from fewer false detections (present structures +0.25 and $-$0.09). On \enh the peripheral bones go the
other way on present structures ($-$0.63, 21 of 30 classes behind), and the heart, vessels and muscles are behind by 0.1--0.6 points
in most classes.

\paragraph{Normalized surface Dice.} Table~\ref{tab:onemodel_nsd} repeats Tab.~2 of the main paper with the normalized surface
Dice (NSD, Sec.~5.2 of the main paper) in place of \Dsc, at 3\,mm, the tolerance of TotalSegmentator's own evaluation, on the
same predictions and with the same rules and macro. The NSD of our ResEnc U-Net differs from the specialists' as its \Dsc does: on \TS it
is ahead on present structures (+0.44 points, 0.18--0.75) and within the interval FP-penalized (+0.71), on \enh it is ahead
FP-penalized (+0.57, 0.33--0.83) and level on present structures (+0.02). At 1 and 2\,mm the same intervals exclude 0, and the differences largely agree in sign
(\TS +0.53 and +0.75 FP-penalized, +0.43 and +0.52 on present structures; \enh +0.40 and +0.52,
$-$0.13 and $-$0.01). The NSD of the plain U-Net follows its \Dsc as well. On \TS it is within the intervals under both rules
at every tolerance. On \enh it is behind on present structures, by 0.27 points at 3\,mm (0.09--0.48) and by 0.63 and 0.38 at 1 and
2\,mm, and within the interval FP-penalized at 2 and 3\,mm ($-$0.17 and $-$0.07), while at 1\,mm it is behind there too,
by 0.40 (0.12--0.69). The \TS data are stored at 1.5\,mm, where a tolerance of 1\,mm accepts only boundary points that match
exactly.

\begin{table}[h]
\caption{\textbf{One model vs.\ the composed specialists in normalized surface Dice} (NSD at 3\,mm), Tab.~2 of the main paper
with NSD in place of \Dsc: the same predictions, test splits (109 / 337 scans), classes (117 / 125) and rules, mean over all
classes. $\Delta$: difference to the specialists; $^*$: its 95\,\% paired case-bootstrap interval excludes 0.}
\label{tab:onemodel_nsd}
\centering
\footnotesize
\setlength{\tabcolsep}{5pt}
\begin{tabular}{@{}lcccc@{}}
\toprule
 & \multicolumn{2}{c}{FP-penalized} & \multicolumn{2}{c}{present structures} \\
\cmidrule(lr){2-3}\cmidrule(l){4-5}
 & NSD & $\Delta$ & NSD & $\Delta$ \\
\midrule
\multicolumn{5}{@{}l}{\textit{\TS labels (117 classes)}} \\
\quad 5 specialists & 93.51 & & 95.53 & \\
\quad 1 plain U-Net & 93.81 & +0.29 & 95.68 & +0.15 \\
\quad 1 ResEnc U-Net & 94.22 & +0.71 & 95.97 & +0.44$^*$ \\
\multicolumn{5}{@{}l}{\textit{\enh labels (125 classes)}} \\
\quad 6 specialists & 93.73 & & 96.57 & \\
\quad 1 plain U-Net & 93.66 & $-$0.07 & 96.30 & $-$0.27$^*$ \\
\quad 1 ResEnc U-Net & 94.29 & +0.57$^*$ & 96.60 & +0.02 \\
\bottomrule
\end{tabular}
\end{table}

\begin{table}[h]
\caption{\textbf{One model vs.\ the composed specialists, per class, summarized by specialist} (test splits of Tab.~2 of the main
paper; one training run per arm). specialist: the class set of one specialist network, named as in the toolkits (``cardiac'' includes
the large vessels; \TS's ``muscles'' also the large bones and the brain). share: median over the classes of the share
of a class's reference boundary that touches a class of another specialist. $\Delta$: mean over the classes of one model $-$
specialists, \Dsc points; row ``all'' is the $\Delta$ of Tab.~2. ahead / behind: classes whose 95\,\% paired case-bootstrap interval
of $\Delta$ lies above / below 0; about 5\,\% of the classes would fall outside by chance alone.}
\label{tab:perclass}
\centering
\footnotesize
\setlength{\tabcolsep}{5pt}
\begin{tabular}{@{}llrrrrrrrr@{}}
\toprule
 & & & & \multicolumn{3}{c}{FP-penalized} & \multicolumn{3}{c}{present structures} \\
\cmidrule(lr){5-7}\cmidrule(l){8-10}
dataset & specialist & classes & share & $\Delta$ & ahead & behind & $\Delta$ & ahead & behind \\
\midrule
\TS & organs & 24 & 0.01 & +1.90 & 4 & 0 & +0.25 & 2 & 0 \\
 & vertebrae & 26 & 0.02 & $-$0.45 & 0 & 3 & $-$0.16 & 0 & 1 \\
 & cardiac & 18 & 0.03 & $-$0.31 & 2 & 1 & +0.68 & 4 & 0 \\
 & muscles & 23 & 0.00 & +1.22 & 2 & 5 & $-$0.09 & 0 & 5 \\
 & ribs & 26 & 0.05 & +0.16 & 7 & 0 & +0.93 & 10 & 0 \\
 & all & 117 & 0.02 & +0.52 & 15 & 9 & +0.31 & 16 & 6 \\
\midrule
\enh & organs & 17 & 0.01 & +0.04 & 1 & 0 & $-$0.02 & 1 & 1 \\
 & vertebrae & 28 & 0.01 & +0.18 & 1 & 7 & $-$0.24 & 0 & 7 \\
 & cardiac & 13 & 0.03 & $-$0.19 & 1 & 12 & $-$0.21 & 1 & 12 \\
 & muscles & 10 & 0.02 & $-$0.29 & 0 & 10 & $-$0.29 & 0 & 10 \\
 & peripheral bones & 30 & 0.00 & +1.04 & 1 & 7 & $-$0.63 & 0 & 21 \\
 & ribs & 27 & 0.04 & +0.89 & 22 & 0 & +0.90 & 22 & 0 \\
 & all & 125 & 0.01 & +0.44 & 26 & 36 & $-$0.06 & 24 & 51 \\
\bottomrule
\end{tabular}
\end{table}

\section{Order-0 label augmentation: paired trainings}
\nnunet 2.8.1's spatial augmentation resamples the label map as one channel per label in the patch, interpolated linearly, and takes
the per-voxel argmax; we resample the label map with nearest neighbor (order 0). To measure how this affects segmentation quality,
we trained each of the 11 specialists of Tab.~2 of the main paper a second time with only the label
resampling changed and scored both against the specialist's own labels on the test splits (Table~\ref{tab:paired}; one run, one
seed per arm). %
On present structures order 0 differs by $-$0.36 to +1.29 \Dsc points per specialist, and by $-$0.40 to +0.50 on the structures a
model detects; the composed specialists differ by +0.14 [$-$0.04, 0.33] (\TS) and +0.09 [$-$0.06, 0.24] (\enh). %
FP-penalized, order 0 is lower by 0.36--0.86 points on four specialists and by 0.32 [0.06, 0.60] on the composed \TS specialists.
Except for the \TS ribs, these differences come from predictions of structures a scan does not contain, which order 0 makes as
often or more often in 10 of the 11 pairs (more in 6 of the 7 pairs that differ, sign test $p = 0.13$); not counting such
predictions under 100 voxels, the composed difference is +0.14 (\TS) and $-$0.04 (\enh). %
The run-to-run spread of the specialists is not measured. The closest reference, three unseeded runs of one ResEnc U-Net on the
same test splits, spreads by up to 0.51 points FP-penalized and 1.3 on present structures within one group of \TS classes, and the
\enh peripheral bones by an s.d.\ of 1.0 FP-penalized; it is a different model and budget, so it is only indicative. We read the
pairs as similar segmentation quality, with a possible shift toward more false detections on absent structures that one seed per
arm cannot separate from noise. %
The CPU cost differs by an order of magnitude. Per training sample (one thread, \TS 192$^3$ patches with the labels merged into
$C$ classes, mean of 30 batches on an L40S node), order 0 takes 0.41--0.58\,s at every $C$ from 1 to 117 (0.49\,s at 117 classes),
while the label-channel resampling takes 4.97\,s at 117 classes in our code and \nnunet 2.8.1's whole augmentation 4.57\,s; its cost
grows with the labels present in the patch, so the medians are lower (2.97 and 2.62\,s). To keep one L40S busy at 192$^3$ and batch
2 with 117 classes, \nnunet 2.8.1 needs about 13 of the 28 augmentation workers a node gives each GPU, order 0 about 1.4. %

\begin{table}[h]
\caption{\textbf{Order-0 label augmentation, paired trainings.} Each specialist network trained twice, with \nnunet 2.8.1's label
resampling (stock) and with nearest neighbor (order 0), one run each; $\Delta$ = order 0 $-$ stock in \Dsc points on the test split,
FP-penalized, on present structures, and on the (scan, structure) pairs the model detects. Absent-structure predictions: (scan,
structure) pairs predicted although the reference does not contain the structure. Composed: the specialists merged as the toolkits
merge them, with 95\,\% paired case-bootstrap intervals.}
\label{tab:paired}
\centering
\footnotesize
\setlength{\tabcolsep}{4pt}
\begin{tabular}{@{}lrrrrr@{}}
\toprule
specialist (classes) & $\Delta$ FP-pen. & $\Delta$ present & $\Delta$ detected & \multicolumn{2}{c}{absent-structure predictions} \\
\cmidrule(l){5-6}
 & & & & stock & order 0 \\
\midrule
\multicolumn{6}{@{}l}{\TS, 109 test scans, 1000 epochs} \\
organs (24) & $-$0.43 & $-$0.02 & $-$0.40 & 76 & 85 \\
vertebrae (26) & 0.00 & +0.12 & +0.07 & 20 & 22 \\
cardiac (18) & $-$0.86 & +1.29 & +0.50 & 13 & 42 \\
muscles (23) & $-$0.05 & $-$0.02 & $-$0.03 & 25 & 26 \\
ribs (26) & $-$0.36 & $-$0.36 & $-$0.26 & 15 & 15 \\
composed (117) & $-$0.32 [$-$0.60, $-$0.06] & +0.14 [$-$0.04, 0.33] & $-$0.05 & 146 & 189 \\
\midrule
\multicolumn{6}{@{}l}{MOOSE on \enh, 337 test scans, 2000 epochs} \\
organs (17) & +0.07 & +0.05 & +0.09 & 36 & 35 \\
vertebrae (28) & +0.06 & +0.31 & $-$0.14 & 54 & 65 \\
cardiac (13) & +0.04 & +0.04 & +0.04 & 6 & 6 \\
muscles (10) & $-$0.02 & $-$0.02 & $-$0.02 & 0 & 0 \\
ribs (27) & +0.10 & +0.10 & +0.10 & 13 & 13 \\
peripheral bones (30) & $-$0.76 & $-$0.05 & $-$0.05 & 358 & 384 \\
composed (125) & $-$0.14 [$-$0.31, 0.04] & +0.09 [$-$0.06, 0.24] & $-$0.01 & 467 & 503 \\
\bottomrule
\end{tabular}
\end{table}

\section{Datasets and licenses}
All data are public and de-identified, and we collected none ourselves; consent for the original acquisitions lies with the
dataset providers. TotalSegmentator v3.0.0 (1,939 CT scans; Zenodo record
22688904) is released under CC BY 4.0. \enh (ScienceDB,
doi:10.57760/sciencedb.34150) is licensed per contributing site: CC BY-NC 4.0 for the cases from the AutoPET challenge, CC BY 4.0
for the others. AMOS 2022 is released under CC BY 4.0 (Zenodo record 7262581). BTCV (Synapse syn3193805) states no license; we use
it as distributed for research. As a secondary use of public, de-identified data, this work required no additional ethics review.

\end{document}